\documentclass[conference, compsoc]{IEEEtran}
\IEEEoverridecommandlockouts

\usepackage{cite, hyperref, tikz, orcidlink}
\usepackage{textcomp}
\usepackage{xcolor}
\usepackage[misc]{ifsym}
\usepackage[fixed]{fontawesome5}

\usepackage{multirow}
\usepackage{graphicx, lipsum}
\usepackage{mathtools}
\usepackage{amsmath,amssymb,amsfonts}
\usepackage{subcaption}
\usepackage[flushleft]{threeparttable}

\usepackage{algorithm}
\usepackage{algpseudocode}
\algnewcommand\algorithmicinput{\textbf{Input:}}
\algnewcommand\Input{\item[\algorithmicinput]}
\algnewcommand\algorithmicoutput{\textbf{Output:}}
\algnewcommand\Output{\item[\algorithmicoutput]}

\DeclarePairedDelimiter{\norm}{\lVert}{\rVert}

\newcommand{\df}[1]{\emph{AGSD}}
\newcommand{\dfh}[1]{\emph{AGSD (ID)}}
\newcommand{\dfo}[1]{\emph{AGSD (OOD)}}
\newcommand{\dfg}[1]{\emph{AGSD (Noise)}}
\newcommand{\sota}[1]{SOTA}

\newcommand{\changed}[1]{\textcolor{black}{#1}}

\newcommand{\take}[1]{\textit{#1}}
\newcommand{\settings}[1]{(Settings: #1)}

\newcommand\copyrighttext{%
  \footnotesize \textcopyright 2024 IEEE. To appear in the proceedings of the 40th Annual Computer Security Applications Conference (ACSAC 2024).
  This is the author’s version of the work. It is posted here for your personal use. Not for redistribution. 
  }
\newcommand\copyrightnotice{%
\begin{tikzpicture}[remember picture,overlay]
\node[anchor=south,yshift=30pt] at (current page.south) {\fbox{\parbox{\dimexpr0.75\textwidth-\fboxsep-\fboxrule\relax}{\copyrighttext}}};
\end{tikzpicture}%
}

\def\BibTeX{{\rm B\kern-.05em{\sc i\kern-.025em b}\kern-.08em
    T\kern-.1667em\lower.7ex\hbox{E}\kern-.125emX}}
\begin{document}

\title{Adversarially Guided Stateful Defense Against Backdoor Attacks in Federated Deep Learning}

\author{
    \IEEEauthorblockN{
        Hassan Ali\orcidlink{0000-0002-1701-0390}
    }
        \IEEEauthorblockA{
            \textit{UNSW Sydney} \\
            {hassan.ali@unsw.edu.au}
        }
    \and
    \IEEEauthorblockN{
        Surya Nepal\orcidlink{0000-0002-3289-6599}
    }
        \IEEEauthorblockA{
            \textit{Data61 CSIRO}\\
            {surya.nepal@data61.csiro.au}
        }
    \and
    \IEEEauthorblockN{
        Salil S. Kanhere\orcidlink{0000-0002-1835-3475} 
    }
        \IEEEauthorblockA{
            \textit{Data61 CSIRO}\\
            {salil.kanhere@unsw.edu.au}
        }
    \and
        \IEEEauthorblockN{
            Sanjay Jha\orcidlink{0000-0002-1844-1520} 
        }
        \IEEEauthorblockA{
            \textit{Data61 CSIRO}\\
            {sanjay.jha@unsw.edu.au}
        }
}
\maketitle

\begin{abstract}
Recent works have shown that Federated Learning~(FL) is vulnerable to backdoor attacks. Existing defenses cluster submitted updates from clients and select the best cluster for aggregation. However, they often rely on unrealistic assumptions regarding client submissions and sampled clients population while choosing the best cluster. We show that in realistic FL settings, state-of-the-art (\sota{}) defenses struggle to perform well against backdoor attacks in FL. To address this, we highlight that backdoored submissions are adversarially \textit{biased} and \textit{overconfident} compared to clean submissions. We, therefore, propose an Adversarially Guided Stateful Defense (\df{}) against backdoor attacks on Deep Neural Networks (DNNs) in FL scenarios. \df{} employs adversarial perturbations to a small held-out dataset to compute a novel metric, called the trust index, that guides the cluster selection without relying on any unrealistic assumptions regarding client submissions. Moreover, \df{} maintains a trust state history of each client that adaptively penalizes backdoored clients and rewards clean clients. In realistic FL settings, where \sota{} defenses mostly fail to resist attacks, \df{} mostly outperforms all \sota{} defenses with minimal drop in clean accuracy (5\% in the worst-case compared to best accuracy) even when (a) given a very small held-out dataset---typically \df{} assumes 50 samples ($\leq 0.1\%$ of the training data) and (b) no held-out dataset is available, and out-of-distribution data is used instead. For reproducibility, our code will be openly available at: \href{https://github.com/hassanalikhatim/AGSD}{https://github.com/hassanalikhatim/AGSD}.
\end{abstract}


\copyrightnotice
\section{Introduction}
Federated Learning~(FL) allows several private data holders (also known as clients) to train a Deep Neural Network~(DNN) on the central server without requiring the server to have access to the clients' data.
Due to its privacy-preserving aspect, FL is applied to several real-world scenarios where security is a major concern, such as autonomous vehicles~\cite{pokhrel2020federated}, healthcare~\cite{rieke2020future, chen2019communication} and IoT devices~\cite{hard2018federated, chen2020asynchronous}. The success of FL is evidenced by companies such as Apple and Google using it to develop products and services for customers~\cite{mcmahan2017communication, paulik2021federated, xie2022federatedscope}.
However, FL requires a certain degree of trust between the server and the clients, which may be abused by either party, allowing several vulnerabilities~\cite{shokri2017membership, carlini2022membership, boenisch2023federated, tramer2022truth, nguyen2023active, ali2021all}, including backdoor attacks~\cite{gu2019badnets, choquette2021label}.

\noindent
\textbf{Backdoor Attacks:}
A backdoor attack in FL occurs when a malicious client locally poisons a subset of its training data (hidden from the server) and submits DNN updates from the poisoned data~\cite{gu2019badnets, zhang2022neurotoxin, nguyen2023iba, schwarzschild2021just}. The backdoored model acts normally for benign inputs and only malfunctions when the attacker's chosen trigger is present in the input.
Due to their high Attack Success Rates (ASR)\footnote{We define the attack success rate as the ratio of correctly classified samples, not originally belonging to the attacker's target class, classified into the target class after poisoning.} with physically realizable triggers, backdoor attacks are viewed by industrial AI practitioners as one of the most concerning threats to FL~\cite{kumar2020adversarial}.
This paper aims to defend FL against backdoor attacks by \emph{honest-but-malicious clients}: assuming that both the server and the clients conform to the designed protocol (honest), but some of the clients have malicious intentions, as typically assumed by recent backdoor attacks and defenses~\cite{nguyen2023iba, nguyen2022flame, krauss2023mesas, zhang2023flip}.

\begin{figure}
    \centering
    \includegraphics[width=1\linewidth]{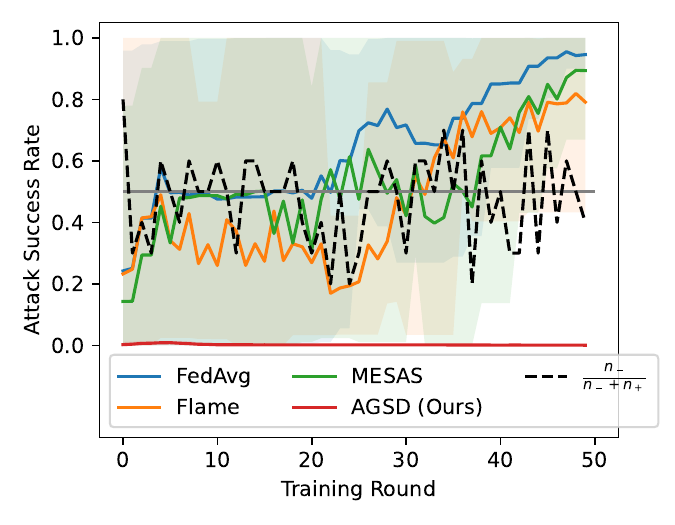}
    \caption{Among randomly sampled clients, malicious clients $n_-$ may outnumber the sampled clean clients $n_+$ in many rounds (e.g., $\frac{n_-}{n_-+n_+} \geq 0.5$ in the figure), invalidating {\sota{}} defenses' assumption~\cite{nguyen2022flame, fung2018mitigating, krauss2023mesas, zhang2023flip}, thereby backdooring the defenses.
    \settings{All settings are similar to MESAS and Flame, except that the clients are sampled randomly}.}
    \label{fig:motivation}
\end{figure}

\noindent
\textbf{Limitations of Backdoor Defenses:}
Many defense mechanisms have been proposed to counter backdoor attacks in FL~\cite{blanchard2017machine, fung2018mitigating, rieger2022deepsight, nguyen2022flame, walter2023optimally, zhang2023flip, krauss2023mesas}. 
Most of them first compute a predefined statistical metric to quantify the similarity among client submissions and cluster these submissions based on the computed metric. Finally, the best cluster is selected based on its proximity to the original DNN~\cite{blanchard2017machine, rieger2022deepsight, fung2018mitigating, nguyen2022flame} or size of the cluster~\cite{nguyen2022flame, krauss2023mesas}. Proximity-based defenses~\cite{blanchard2017machine, rieger2022deepsight, fung2018mitigating, nguyen2022flame} not only hinder DNN training (resulting in poor accuracy) but are also vulnerable to adaptive attacks~\cite{krauss2023mesas, nguyen2023iba}. On the contrary, population-based defenses~\cite{nguyen2022flame, krauss2023mesas} implicitly assume that benign clients outnumber backdoored clients among the clients sampled to update the DNN in each training round.
However, under a realistic FL threat model, when clients are mostly sampled randomly~\cite{chen2019communication, li2020review, chen2020asynchronous, pokhrel2020federated}, this assumption is invalidated at several training rounds. We show that in such realistic situations, all \sota{} defenses evaluated in this paper (including the proximity-based defenses) can be circumvented by backdoor attacks with ~100\% ASR in most cases. The underlying assumption of these defenses is more frequently invalidated when the number of malicious clients is comparable to the number of clean clients in the clients' universal set---an assumption made by most {state of the art} defenses~\cite{nguyen2022flame, krauss2023mesas, zhang2023flip}.
This is illustrated in Fig.~\ref{fig:motivation}, where the backdoor ASR continues to increase gradually, showing abrupt increment at rounds when malicious clients outnumber benign clients ($\frac{n_-}{n_- + n_+} > 0.5$).

\noindent \textbf{Our Work:}
In this paper, we propose a novel method for a server to identify the best cluster to update the DNN irrespective of the cluster size and its proximity to the DNN. We first highlight that backdoored classifiers are adversarially \emph{biased}---when adversarially attacked, backdoored classifiers are notably more inclined towards the backdoor target class as compared to the clean classifiers---and \emph{overconfident}---backdoored classifiers classify adversarial perturbations notably more confidently as compared to the clean classifiers.
These two properties of backdoored classifiers can be leveraged to detect backdoored client submissions during federated training, formulating an Adversarially Guided Stateful Defense (\df{}) against backdoor attacks in FL. 


\noindent \textbf{Approach:}
\df{} works in four stages: (1) In the \textit{preliminary aggregation stage}, \df{} scales and preliminarily aggregates client submissions via federated averaging; (2) In the \textit{clustering stage}, \df{} clusters client submissions using spectral clustering algorithm~\cite{shi2003multiclass} based on their difference from the preliminary aggregated model and the initial model from the previous round; (3) In the \textit{guided selection stage}, \df{} computes adversarial perturbations on the preliminary aggregated model using a small held-out dataset assumed to be available to \df{} and transfers these perturbations to the client submissions to compute a novel metric, called the trust index $\gamma_i$ for each client $c_i$ sampled for the training round. $\gamma_i$ quantifies the non-maliciousness of the client submission. \df{} selects the cluster of clients that exhibits the highest average value of $\gamma_i$; (4) In the \textit{stateful selection stage}, \df{} maintains each client's trust history $\phi_i$ that is adaptively updated based on $\gamma_i$ at each training round. Only clients in the selected cluster with $\phi_i > 0$ can update the model.

\noindent \textbf{Findings and Contributions:}
We evaluate \df{} on three different benchmarks (MNIST, CIFAR-10 and GTSRB) commonly used to compare backdoor defense robustness~\cite{nguyen2022flame, krauss2023mesas}. However, for comprehensive evaluations, we primarily rely on the GTSRB dataset due to its practical relevance to autonomous vehicles-- now being foreseen as one of the major applications of FL in the future~\cite{li2020review, dai2023online, pokhrel2020federated} with countries like the UK and Australia enacting legislation to support its use~\cite{ukpass, aupass}.

\df{} only employs adversarial perturbations to guide the clustering stage instead of adversarially training client submissions or the aggregated model (which negatively affects the main task accuracy~\cite{khalid2019qusecnets}). This lets \df{} outperform \sota{} defenses, even when given a very small held-out dataset, typically comprising only 50 data samples ($\leq 0.1\%$ of the training data), with only slight degradation in the main task accuracy.

We also consider instances where there is no held-out data for \df{}. In such cases, \df{} uses an {out of distribution} (OOD) data as the healing set (e.g. we use the CIFAR-10 dataset for processing Resnet-18 submissions trained on the GTSRB dataset). Interestingly, \df{} outperforms {\sota{}} defenses even in these scenarios. We observe that even when using OOD data as the held-out set, \df{} can successfully identify malicious clients. We conjecture that adversarial bias and overconfidence---the two highlighted properties of backdoored classifiers---are independent of the underlying training data distribution and are instead enabled by the local minimas created by backdoor features. Through extensive evaluation, we show that \df{} is sufficiently robust against changes in hyperparameters and adaptive backdoor attacks.


Our main contributions are summarized below:
\begin{enumerate}
    \item We highlight two properties of the backdoored classifiers---bias and overconfidence---enabled by adversarial perturbations that can be used to detect backdoored client submissions.
    \item We propose a novel metric called the trust index, denoted by $\gamma$, to quantify the non-maliciousness of client submissions based on a small held-out dataset.
    \item We propose an Adversarially Guided Stateful Defense (\df{}) against backdoor attacks in realistic FL settings. Unlike \sota{} defenses, \df{} does not make any assumptions regarding the population of sampled clients and resists backdoor attacks where \sota{} defenses struggle to perform well.
    \item \df{} only minimally affects the main task performance and works well with a very small held-out dataset (typically, we set the size of the held-out dataset to be $\leq$0.1\% the train set size), making it effective for practical purposes.
    \item Even when no held-out dataset is available, \df{} can make use of the {out of distribution} dataset to generate adversarial perturbations and outperform \sota{} defenses on both standard and adaptive backdoor attacks.
\end{enumerate}

\noindent
\changed{
\textbf{Practicality of \df{}:} \df{} needs a very small held-out dataset (typically only ~50 samples) and can work with the OOD dataset. These assumptions, particularly the one assuming access to the OOD dataset, hold for almost all practical scenarios because of several openly available datasets. Additionally, \df{} can resist backdoors even if the OOD dataset has fewer classes, smaller input sizes, and non-overlapping classes than the training dataset as we show later in our experiments. For example, we used CIFAR-10 (32x32x3 images of 10 classes) as the OOD dataset to train Resnet-18 on GTSRB (45x45x3 images of 43 classes).
}

\noindent 
\textbf{Paper Outline:}
Sec~\ref{sec:Methodology} formulates backdoor attacks in FL and describes the working of \df{}, Sec~\ref{sec:evaluation} details our experimental setup, compares our evaluation results with \sota{} defenses on \sota{} backdoor attacks and adaptive attacks, and presents additional results and insights into the effectiveness of \df{}.

\section{Methodology}\label{sec:Methodology}
In this section, we first present the problem formulation of backdoor attacks in FL, highlighting challenges in defending against these attacks and laying the foundations for our defense. We then formally describe the working of \df{}: a novel adversarially guided stateful defense against backdoor attacks in \ref{sec:hgsd_methodology}.

\subsection{Problem Formulation}\label{subsec:methodology_problemFormulation}
We assume a differentiable untrained classifier $f$ and a dataset $D = \{(x_i, y_i)\}_{i=0}^{|D|-1}$ on which $f$ is trained, where $|D|$ denotes the size of $D$. We denote the training process by $f_+ \leftarrow \mathcal{T}(D, f)$ that optimizes $f$ on $D$ using gradient descent to produce $f_+$ as the trained model.

\noindent \textbf{Backdoor attacks:}
Given $D$, backdoor attacks typically work by poisoning randomly selected data samples $B \subset D$ with a trigger $\tau$ and mislabelling the samples to the target class $y_\tau$, where typically $|B|/|D| \leq 3\%$.

\begin{equation}
    D_- = \{(x^i, y^i)\}_{\forall i \notin B} + \{(x^i+\tau, y_\tau)\}_{\forall i \in B}
    \label{eq:backdoor_attack}
\end{equation}

Backdoored classifier $f_- \leftarrow \mathcal{T}(D_-, f)$ achieves a similar main task accuracy as the clean classifier $f_+$ but differs on the inputs poisoned by $\tau$. Formally, $\forall (x,y) \in D$,
\begin{subequations}
    \begin{gather}
        f_-(x) \approx f_+(x) \approx y \label{eq:backdoor_attack_effect_clean} \\
        f_+(x+\tau) \approx y \neq f_-(x+\tau) \approx y_\tau \label{eq:backdoor_attack_effect_poisoned}
    \end{gather}
\end{subequations}

Eq-\eqref{eq:backdoor_attack_effect_poisoned} highlights that backdoor effects can be removed by optimizing the following loss function,
\begin{equation}
    \operatorname{minimize}\ \norm{f_+(x+\tau) - f_-(x+\tau)}
    \label{eq:countering_backdoor_attacks}
\end{equation}
However, the trigger $\tau$ is known only to the attacker, which makes it challenging to detect and counter the attack.

\noindent \textbf{Federated Learning (FL):}
We consider a server training a global classifier $f$ in federated learning setting comprising of $n$ clients $\{C_i\}_{i \in [0,...,n-1]}$, where $C_i$ holds the dataset $D_i$ of size $|D_i|$. 

At each training iteration $t$, the server uses a sampling function $S(c, n)$ to select $c \leq n$ clients randomly, shares the updated global classifier $f_{t-1}$ with selected clients and receives the locally updated classifiers as client submissions $\{f_{t,i} \leftarrow \mathcal{T}(D_i, f_{t-1})\}_{i \in S(c, n)}$, which are aggregated by the server to compute the updated classifier $f_t$.
\begin{equation}
    f_t = \frac{1}{c} \sum_{i \in S(c,n)} f_{t,i} = \frac{1}{c} \sum_{i \in S(c,n)} \mathcal{T}(D_i, f_{t-1})
    \label{eq:federated_averaging}
\end{equation}

\changed{
\noindent \textbf{Threat model:}
Following prior defense~\cite{blanchard2017machine, rieger2022deepsight, fung2018mitigating, nguyen2022flame, krauss2023mesas}, we assume honest-but-malicious clients: the clients stick to the appropriate protocol designed by the software, but $p \leq n$ clients who are controlled by the attacker have malicious intents---they intentionally submit a poisoned classifier $f_{t,i-} \leftarrow \mathcal{T}(D_{i-}, f_{t-1})$ instead of $f_{t,i}$ to the server by poisoning $|B_i| \leq |D_i|$ data samples held by them. We experiment with different $\frac{p}{n}$, but typically assume $\frac{p}{n}$ = 0.45, following previous works~\cite{nguyen2022flame, krauss2023mesas}. Moreover, our attacker can choose to/not-to freely manipulate local model weights of the controlled clients, and can craft adaptive attacks after knowing about AGSD (Section~\ref{sec:adaptive_evaluation}). In the paper, we also refer to these malicious clients as backdoored clients.
}

\begin{figure}
    \centering
    \includegraphics[width=0.7\linewidth]{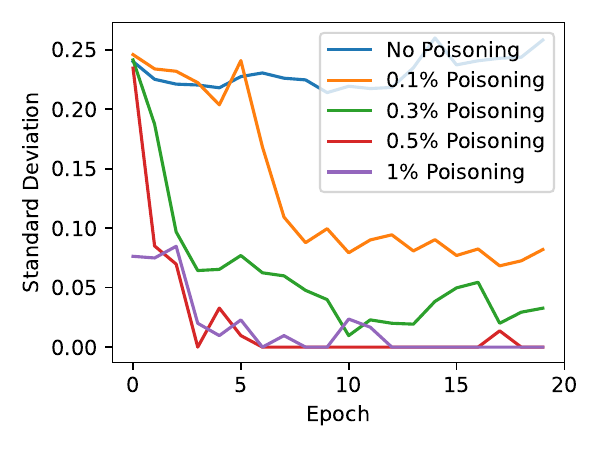}
    \caption{Standard deviation of the output classes of clean and backdoored classifiers for adversarial inputs.}
    \label{fig:std_id}
\end{figure}
\subsection{\df{}: Adversarially Guided Stateful Defense Against Backdoor Attacks}
\label{sec:hgsd_methodology}

\noindent \textbf{Observation 1: Backdoored classifiers are adversarially biased.}
Let $f_+ \leftarrow \mathcal{T}(D, f)$ and $f_- \leftarrow \mathcal{T}(D_-, f)$ be trained classifiers on clean and backdoored datasets, denoted as $D$ and $D_-$ respectively. We note that when an arbitrary input sample $(x, y) \in D$ is perturbed by an untargeted adversarial attack $\mathcal{A}$, the backdoored classifier is highly likely to output the target class $y_t$ on the perturbed sample as compared to the clean classifier.

\begin{equation}
    \mathbb{E}\left[ f_+(x + \mathcal{A}(x)) - y_t \right] > \mathbb{E}\left[ f_-(x + \mathcal{A}(x)) - y_t \right]
    \label{eq:adv_bias}
\end{equation}

To illustrate this, we randomly choose 500 samples from the GTSRB dataset, adversarially perturb them using FGSM attack, and plot the standard deviations of the output classes as predicted by clean and poisoned ResNet-18 classifiers over the perturbed dataset for different epochs in Fig.~\ref{fig:std_id}. Note that the strongest backdoor attack results in the smallest standard deviation in the output classes in Fig.~\ref{fig:std_id}. This is because the stronger the attack, the greater the expectation of $f_-(d+\mathcal{A}(x))$ to be $y_t$.

\begin{figure}
    \centering
    \includegraphics[width=0.7\linewidth, page=2]{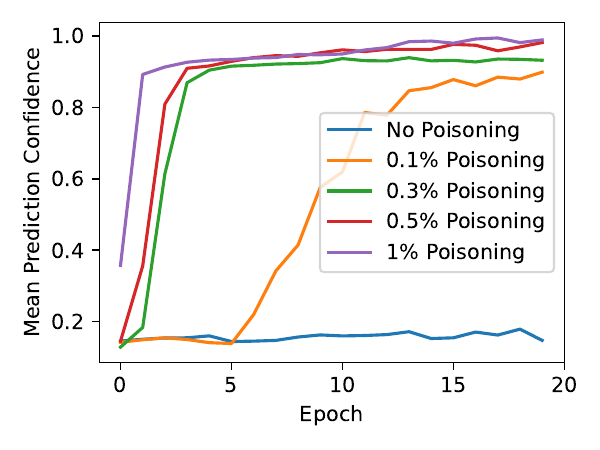}
    \caption{Confidence of clean and backdoored classifiers when classifying adversarial inputs.}
    \label{fig:conf_id}
\end{figure}
\noindent \textbf{Observation 2: Backdoored classifiers are adversarially overconfident.} Considering similar settings as above, the perturbed samples are misclassified by the backdoored classifier with notably higher confidence as compared to the clean classifier. Formally,

\begin{equation}
    \max \mathbb{E}[ f_+(x + \mathcal{A}(x)) ] < \max \mathbb{E}[ f_-(x + \mathcal{A}(x)) ]
    \label{eq:adv_overconfidence}
\end{equation}

We illustrate this in Fig.~\ref{fig:conf_id} by adversarially perturbing 500 randomly chosen samples from the GTSRB dataset. As previously, a stronger backdoor attack results in a higher confidence when predicting adversarially perturbed samples.

\begin{figure*}
    \centering
    \includegraphics[width=0.7\linewidth]{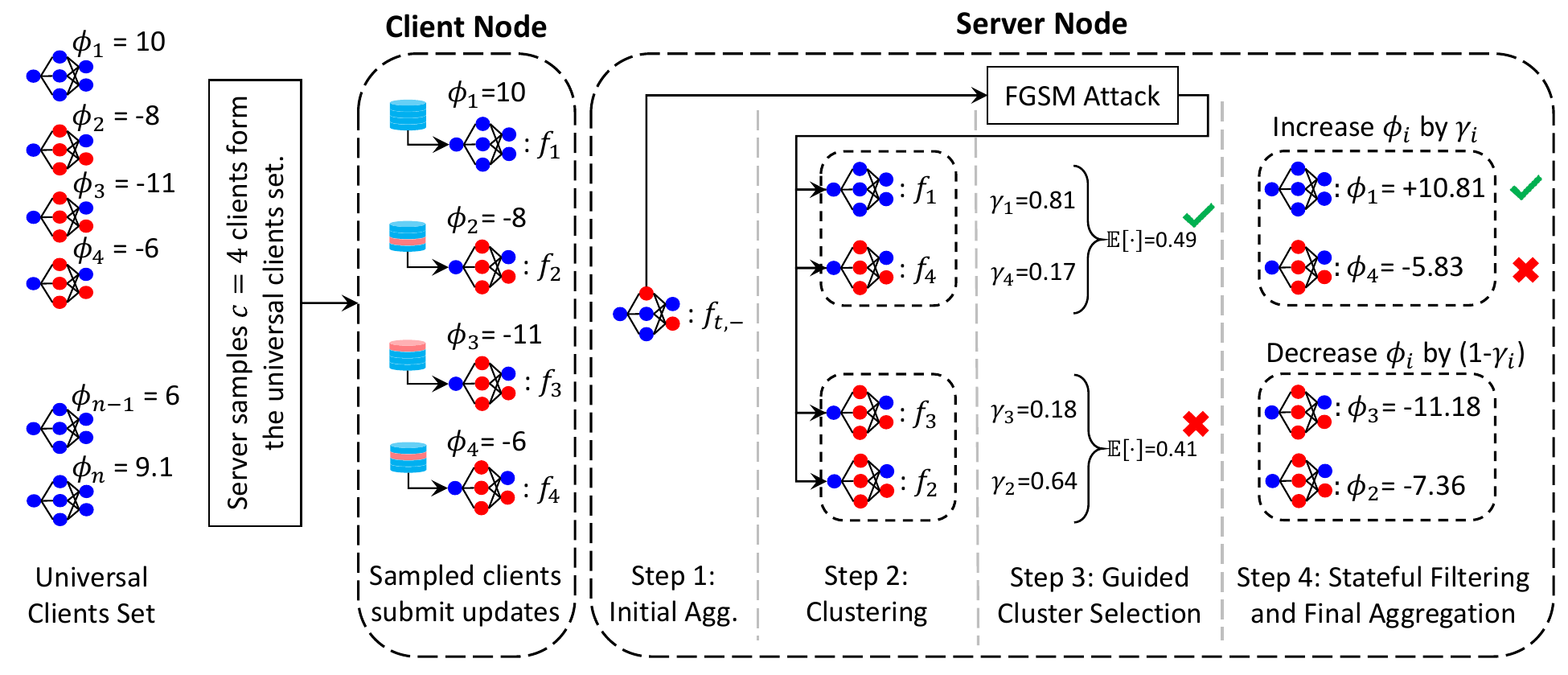}
    \caption{Illustration of the working of \df{} in four steps for training round $t$, where the number of sampled clients $c$ is assumed to be 4. \df{} maintains a trust history $\phi_i$ of each client. After clustering client submissions in Step 2, \df{} uses a novel method to compute the trust-index $\gamma_i$ of each submission in Step 3 to identify the best cluster for the update. In step 4, clients of the best cluster with $\phi_i < 0$ are ruled out of aggregation.}
    \label{fig:methodology}
\end{figure*}
\subsubsection{Methodology}
Let us assume a server that owns a held-out dataset $D_h$. For iteration $t$, the server samples clients $S(c, n)$ comprising of both the clean clients $C_+$ and the backdoored clients $C_-$, who submit the updated models $\{f_{t,i}\}_{\forall i \in S(c, n)} = \{f_{t,i+}\}_{\forall i \in C_+} + \{f_{t,i-}\}_{\forall i \in C_-}$, respectively. \df{} algorithm is given in Alg.~\ref{alg:defense} (Appendix) and illustrated in Fig.~\ref{fig:methodology}. \df{} works in four stages:

\noindent \textbf{(Step 1) Preliminary Aggregation:}
\df{} first receives the {client submissions} $\{f_{t,i}\}_{\forall i \in S(c, n)}$ and computes the difference of each client submission from the aggregated classifier of previous round $f_{t-1}$.
\begin{equation}
    \Delta = \{\delta_{t,i}\}_{\forall i \in S(c, n)} = \{f_{t,i}-f_{t-1}\}_{\forall i \in S(c, n)}
\end{equation}

\df{} then scales all the differences onto an $l_2$ sphere defined by the median of all the $l_2$ differences of the client submissions from $f_{t-1}$.
\begin{multline}
    \Delta^{(s)} = \left\{\delta^{(s)}_{t,i}\right\}_{\forall i \in S(c, n)} = \\
    \left\{\frac{\delta_{t,i} \times \operatorname{median}\norm{\Delta}_2}{ \norm{\delta_{t,i}}_2}\right\}_{\forall i \in S(c, n)}
\end{multline}

The median of the differences has been formally shown to be more robust to outliers as compared to other metrics---such as mean, min and max---in FL~\cite{nguyen2022flame}. \df{} then preliminarily aggregates the scaled differences with $f_{t-1}$ using federated averaging to get the preliminary aggregated classifier $f_{t,-}$,
\begin{equation}
    f_{t,-} = f_{t-1} + \frac{1}{c} \left( \sum_{\forall i \in C_+}{\delta^{(s)}_{t,i+}} + \sum_{\forall i \in C_-}{\delta^{(s)}_{t,i-}} \right)
    \label{eq:preliminary_aggregation}
\end{equation}

Note that $f_{t,-}$ is potentially poisoned because eq-\eqref{eq:preliminary_aggregation} has the same effect as training $f_{t-1}$ on the dataset $\{D_i\}_{\forall i \in C_+} + \{D_{i-}\}_{\forall i \in C_-}$, which is the backdoor attack as defined in eq-\eqref{eq:backdoor_attack}.

\begin{figure}
    \centering
    \includegraphics[width=1\linewidth]{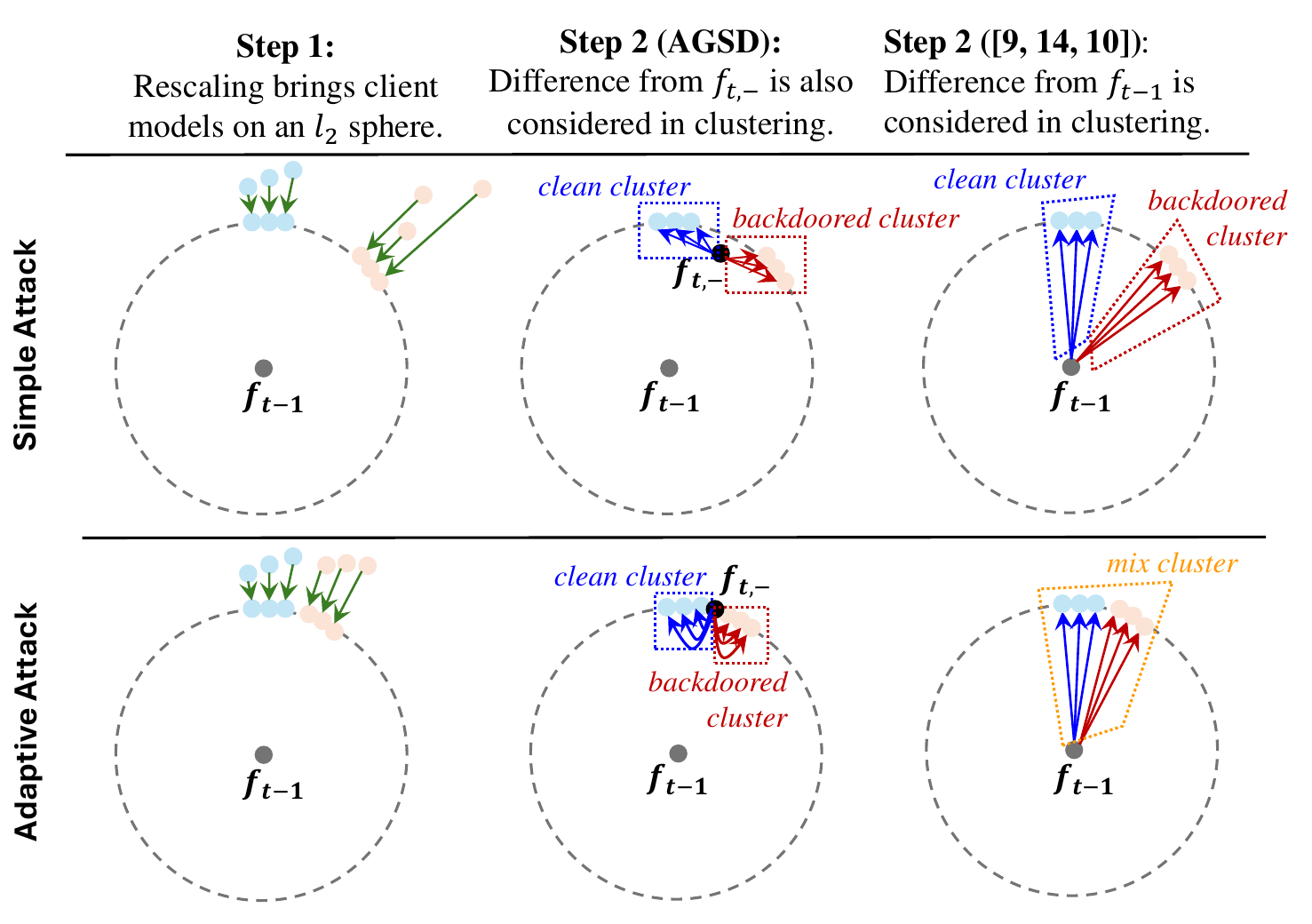}
    \caption{\df{} uses an improved clustering metric based on the difference from the preliminary aggregated model. This lets \df{} distinguish adaptive clients, unlike \sota{} defense.}
    \label{fig:better_clustering_metric}
\end{figure}
\noindent \textbf{(Step 2) Clustering:}
Because the client submissions comprise of both the clean and the backdoored submissions, $f_{t,-}$ (eq-\eqref{eq:preliminary_aggregation}) is also backdoored. To counter this, clustering is used to distinguish between the clean and the backdoored submissions. 

\noindent \textit{\underline{Clustering metrics:}}
Existing \sota{} defenses rely on: (1) the submitted weights~\cite{fung2018mitigating}; or (2) the difference from $f_{t-1}$~\cite{nguyen2022flame, krauss2023mesas} to cluster {client submissions}. However, both these metrics can be bypassed by adaptive attacks~\cite{nguyen2023iba}, that bring backdoored submissions close to the clean submissions (see Fig.~\ref{fig:better_clustering_metric} for illustration). To mitigate this, \df{} improves the statistical metric by also considering the difference from the preliminary aggregated classifier $f_{t,-}$, in addition to the difference from $f_{t-1}$ while clustering client submissions, as illustrated in Fig.~\ref{fig:better_clustering_metric}. Initially, rescaling projects client submissions onto an $l_2$ sphere from $f_{t-1}$. Therefore, the preliminary aggregated classifier $f_{t,-}$ also always lies within the $l_2$ norm of $f_{t-1}$, where $l_2$ norm is determined by the median of $l_2$ norms of submitted client updates.

\noindent \textit{\underline{Clustering algorithm:}}
\df{} uses multi-class spectral clustering~\cite{shi2003multiclass} to cluster submissions into two clusters, $K_1$ and $K_2$ based on the new statistical metric. Spectral clustering is highly effective for clustering DNN updates because of the non-convex clusters with varying variances~\cite{von2007tutorial, damle2019simple}.

\noindent \textbf{(Step 3) Guided Cluster Selection:}
Assuming that benign clients outnumber backdoored clients, prior works identify the largest cluster as the one comprising only clean clients. However, in real-world scenarios, random sampling of clients invalidates the aforementioned observation at several training rounds, leading to a gradual backdooring of the classifier, as observable in Fig.~\ref{fig:motivation}. To overcome this limitation of prior works, \df{} uses a novel method that uses a held-out dataset $D_h$, assumed to be available to \df{}, to first generate adversarial perturbations over $D_h$ and then identify the right cluster based on standard deviations and prediction confidences of the client submissions.

\noindent \textit{\underline{Generating adversarial perturbations:}}
\df{} utilizes single step FGSM attack to optimize a novel loss function on $f_{t,-}$ over $D_h$, as defined below,

\begin{multline}
    \underbrace{\underset{\forall (x,y) \in D_h}{\mathbb{E}}\left[ -\mathcal{L}_c\left( f_{t,-}\left( x+\mathcal{A}(x) \right), y \right) \right]}_\text{$\mathcal{L}_1$: untargeted adv loss} + \\
    \underbrace{f_{t,-}\left( x+\mathcal{A}(x) \right) - \underset{\forall (x,y) \in D_h}{\mathbb{E}}\left[ f_{t,-}\left( x+\mathcal{A}(x) \right) \right]}_\text{$\mathcal{L}_2$: backdoor loss}
    \label{eq:agsd_loss}
\end{multline}

where $\mathcal{L}_c$ denotes the crossentropy loss. 
\changed{
FGSM attack is efficient---uses a single gradient step---and transfers effectively to other classifiers~\cite{ali2019sscnets}, making it a good choice for our defense.
}
$\mathcal{L}_1$ in eq-\eqref{eq:agsd_loss} is the conventionally used untargeted adversarial attack loss that optimizes perturbations $\mathcal{A}(D_h)$ such that $f_{t,-}$ outputs the incorrect class, while $\mathcal{L}_2$ optimizes $\mathcal{A}(D_h)$ such that $f_{t,-}$ outputs the same class for different perturbed data samples. $\mathcal{L}_2$ makes the adversarial attack specifically effective against backdoored submissions because of its similarity to eq-\eqref{eq:backdoor_attack}.

\noindent \textit{\underline{Bias and overconfidence:}}
\df{} then uses the adversarially perturbed samples $D_{adv} = D_h + \mathcal{A}(D_h)$ computed on $f_{t,-}$ to perform transfer attack on the client submissions $\{f_{t,i}\}_{\forall i \in S(c, n)}$ and computes standard deviations $\sigma_i$ and prediction confidences $\eta_i$ of $f_{t,i}(D_{adv})$.

\begin{subequations}
    \begin{gather}
        \sigma_i = \underset{D_{adv}}{\operatorname{std}}( \operatorname{onehot} \operatorname{argmax} f_{t,i}(D_{adv}) ) )
        \label{eq:sigma_i} \\
        \eta_i = \max \underset{D_{adv}}{\mathbb{E}} \left[ f_{t,i}(D_{adv}) \right]
        \label{eq:conf_i}
    \end{gather}
\end{subequations}

Note that instead of individually attacking each client submission, \df{} computes $D_{adv}$ on $f_{t,-}$ and transfers $D_{adv}$ to client submissions, which does not incur significantly high computational costs as compared to the baseline scenario (no defense). All $\sigma_i$ and $\eta_i$ values are collected in two arrays---$\sigma = \{\sigma_i\}_{\forall i \in S(c,n)}$ and $\eta = \{\eta_i\}_{\forall i \in S(c,n)}$---and normalized using the softmax function to get the probability of each client submission to be benign.

\noindent \textit{\underline{Quantifying non-maliciousness:}}
The trust index array $\gamma$ quantifying the trustworthiness of each client is then computed as follows,

\begin{equation}
    \gamma = \{\gamma_i\}_{\forall i \in S(c,n)} = \underset{i}{\operatorname{softmax}}(\sigma) - e^{-\mathcal{W}\left( \sigma \right)} \times \underset{i}{\operatorname{softmax}}(\eta)
    \label{eq:trust_index}
\end{equation}

where $\mathcal{W}(\cdot)$ is defined as,
\begin{equation}
    \mathcal{W}(\sigma) = \frac{\max_i \operatorname{softmax} \sigma - \min_i \operatorname{softmax} \sigma}{\operatorname{mean}_i \operatorname{softmax} \sigma - \min_i \operatorname{softmax} \sigma}
    \label{eq:weight_of_sigma}
\end{equation}

Eq-\eqref{eq:trust_index} will result in a large value of $\gamma$ for clients that have larger standard deviations in the output classes (Fig.~\ref{fig:std_id}) and smaller confidences (Fig.~\ref{fig:conf_id}) when predicting $D_h + \mathcal{A}(D_h)$. The scaling factor $e^{-\mathcal{W}(\sigma)}$ adaptively adjusts the effect of $\eta_i$---when all the $\sigma_i$ values are approximately the same, $\mathcal{W}(\sigma) \approx 0$ and \df{} puts higher trust in $\eta_i$ values. 

\noindent \textit{\underline{Cluster selection:}}
\df{} then selects the cluster of clients that exhibits the greatest average value of $\gamma_i$ as potential candidates to update the model.
\begin{equation}
    \alpha = \underset{k \in \{1,2\}}{\operatorname{argmax}} \left\{ \gamma_k \right\}, \beta = \underset{k \in \{1,2\}}{\operatorname{argmin}} \left\{ \gamma_k \right\}
    \label{eq:best_cluster_selection}
\end{equation}

where $\gamma_k = \underset{\forall i \in K_k}{\mathbb{E}}(\gamma_i)$ is the average value of $\gamma_i$ in $K_k$. The cluster $K_\alpha$ is therefore selected to update the model. 

As we show later in Section~\ref{sec:evaluation}, \df{} is effective even when the size of $D_h$ is less than $0.02\%$ the size of training data, or when $D_h$ is an {out of distribution} dataset.

\noindent \textbf{(Step 4) Stateful Filtering:}
Despite the effectiveness of clustering mechanisms, it is not uncommon for some backdoored client submissions to be clustered together with the clean client submissions~\cite{nguyen2022flame}. To overcome this, \df{} maintains a trust history $\phi_i$ of each client $C_i$, which is \textit{adaptively updated} based on $\gamma_i$ at each training round. A client in $K_\alpha$ is only considered for an update if it exhibits $\phi_i > 0$. 

The updated global model $f_t$ is therefore,
\begin{equation}
    f_t = \frac{1}{|K_\alpha|} \left( \sum_{\forall i \in K_\alpha, \phi_i > 0}{f_{t,i} \times \frac{\min{\norm{\{f_{t,j}\}_{j \in K_\alpha}}}}{\norm{f_{t,i}}} } \right)
    \label{eq:federated_avg_selected}
\end{equation}

The adaptive method updates $\phi_i$ of each client $c_i$.
\begin{equation}
    \phi_i = 
    \begin{cases}
        \phi_i + \frac{\gamma_i}{\max \gamma} & C_i \in K_\alpha \\
        \phi_i - \left( 1 - \frac{\gamma_i}{\max \gamma} \right), & C_i \in K_\beta
    \end{cases}
    \label{eq:update_trust_history}
\end{equation}

The condition in eq-\eqref{eq:update_trust_history} tackles the case when all the sampled clients are benign ($|C_+| = c$). In such cases, $\frac{\gamma_i}{\max \gamma}$ of the rejected clusters is close to 1, resulting in a negligible effect on rejected clients' trust history.
\color{black}

\begin{figure}
    \centering
    \begin{subfigure}{1\linewidth}
        \centering
        \includegraphics[width=0.7\linewidth, page=1]{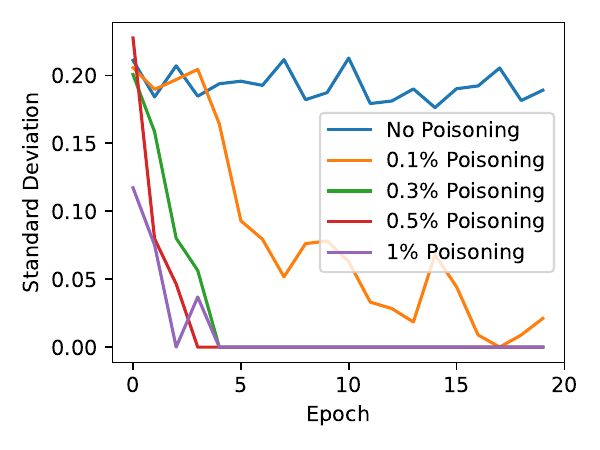}
        \caption{Standard Deviation}
    \end{subfigure}
    \begin{subfigure}{1\linewidth}
        \centering
        \includegraphics[width=0.7\linewidth, page=2]{figures_agsd/fgsm_std_ood_mnist.pdf}
        \caption{Prediction Confidence}
    \end{subfigure}
    \caption{Standard deviation and confidence of clean and backdoored classifiers' outputs when classifying adversarial inputs.}
    \label{fig:std_conf_ood}
\end{figure}
\subsection{Different Modes of \df{}} \label{sec:Methodology_DifferentModes}
\df{} assumes access to a correctly labelled held-out dataset $D_h$. In practice, data annotation requires human expertise and labor, making it challenging and expensive to own a large dataset~\cite{welinder2010online}. Can we instead use an {out of distribution} (OOD) data to adversarially attack and analyze client submissions? To answer this, we repeat the experiment of Fig.~\ref{fig:std_id} and Fig.~\ref{fig:conf_id} with OOD data---we perturb 500 randomly sampled CIFAR-10 samples to adversarially attack two ResNet-18 classifiers being trained on clean and poisoned GTSRB datasets respectively.

As illustrated in Fig.~\ref{fig:std_conf_ood}(a) and (b), even when using OOD data as $D_h$, standard deviations and confidences of output classes show similar trends as those shown when $D_h$ is {in distribution} (ID) data, and clean classifier is easily distinguishable from the backdoored classifiers. As previously observed for ID data in Fig.~\ref{fig:std_id} and Fig.~\ref{fig:conf_id}, a stronger backdoor attack results in a smaller standard deviation and greater confidence in the output classes when predicting adversarially perturbed inputs.
To differentiate between the two modes in our evaluations, we use \dfh{} to refer to \df{} that uses an (ID) held-out data and \dfo{} to denote \df{} that uses OOD held-out data. When using vanilla \df{} in the paper, we are referring to both incarnations of \df{}.

\section{Results and Discussions}\label{sec:evaluation}

This section first details our experimental setup, compares the efficacy of \df{} (in both its incarnations---\dfh{} and \dfo{}) with \sota{} defenses and analyzes the robustness of \df{} to changes in the hyperparameters. We design our experimental setup to answer the following questions:
\begin{enumerate}
    \item How does \df{} compare with the \sota{} backdoor defenses in realistic FL settings?
    \item To what extent is \df{} robust to changes in FL hyperparameters?
    \item How is \df{} affected by data distribution among clients (IID, non-IID)?
    \item How robust is \df{} to the adaptive backdoor attacks as compared to \sota{} backdoor defenses?
\end{enumerate}

\subsection{Experimental Setup} \label{subsec:evaluation_experimental_setup}

\noindent \textbf{Datasets and Model architectures:}
We perform our evaluations on MNIST (10 classes), CIFAR-10 (10 classes) and GTSRB (43 classes) datasets. We train a simple CNN classifier on MNIST with a patience of 50 training rounds and a ResNet-18 classifier on CIFAR-10 and GTSRB datasets with a patience of 150 and 100 training rounds. If the accuracy of the classifier on the test set does not increase for the set number of patience rounds, the server stops the training and restores the best parameters.

\noindent \textbf{Default FL configurations:}
Unless otherwise stated, we use the following FL setting. We divide the given training set among $n = 100$ clients and sample $\frac{c}{n} = 0.1$ clients in each FL round. Following Krauss et. al~\cite{krauss2023mesas}, we set the ratio of malicious clients $\frac{p}{n} = 0.45$ and $|B_i|/|D_i|=0.25$ for each malicious client (25\% data poisoned by each client).
\changed{
Following prior defenses~\cite{nguyen2022flame, krauss2023mesas, rieger2022deepsight}, we typically set the number of clusters to 2. However, we analyze the effect of choosing the number of clusters to be $>2$.
}

Following recent works~\cite{zhang2023flip, krauss2023mesas}, we use 200 training rounds for MNIST and 500 training rounds for CIFAR-10 and GTSRB datasets. We use an SGD optimizer with a learning rate of 0.1, momentum of 0.9 and weight decay of 0.0005.

\noindent \textbf{Default \df{} configurations:}
For \df{} server training, unless otherwise stated, we use the held-out dataset size $|D_h| = 50$, which is $\leq 0.1\%$ of the total training data for all the considered datasets in our experiments.

\noindent \textbf{Backdoor attacks:}
Similar to the related works, we experiment with four different backdoor attacks well-suited for FL scenarios---Visible Trigger Backdoor Attack (VTBA)~\cite{gu2019badnets}, Invisible Trigger Backdoor Attack (ITBA)~\cite{chen2017targeted}, Neurotoxin Backdoor Attack (NBA)~\cite{zhang2022neurotoxin}, and Irreversible Backdoor Attack (IBA)~\cite{nguyen2023iba}. IBA is the most recent \textit{adaptive backdoor attack} that leverages the power of adversarial perturbations to regulate the deviation of backdoored classifiers from the clean classifiers.

\noindent \textbf{Backdoor defenses:}
We compare our proposed defense with several {\sota{}} backdoor defenses---Differentially-Private SGD~\cite{dwork2006calibrating, dwork2014algorithmic}, Foolsgold~\cite{blanchard2017machine}, Mutli-Krum ({\changed{m-Krum}})~\cite{fung2018mitigating}, DeepSight~\cite{rieger2022deepsight}, Flame~\cite{nguyen2022flame} and Mesas~\cite{krauss2023mesas}---that we choose based on their popularity and recency.

\noindent \textbf{Evaluation Metrics:}
Following other works in literature, we use two commonly used metrics to evaluate and compare different models---Accuracy on Clean Data (CA) and Attack Success Rate (ASR).

\begin{table}[t]
\centering
\caption{CA$\uparrow$(ASR$\downarrow$) of FL servers for MNIST dataset.}
\resizebox{\linewidth}{!}{%
\begin{tabular}{c|c|c|c|c|c|}

& \textbf{No Attack} & \textbf{VTBA}~\cite{gu2019badnets} & \textbf{ITBA}~\cite{chen2017targeted} & \textbf{NBA}~\cite{zhang2022neurotoxin} & \textbf{IBA}~\cite{nguyen2023iba} \\
\hline

FedAvg & 0.99(-) & 0.99(1.00) & 0.98(1.00) & 0.79(0.13) & 0.99(1.00)
\\
DP-SGD & 0.99(-) & 0.98(1.00) & 0.99(1.00) & 0.94(0.99) & 0.99(1.00)
\\
{\changed{m-Krum}} & 0.99(-) & 0.99(1.00) & 0.99(1.00) & 0.98(1.00) & 0.99(1.00)
\\
FoolsGold & 0.99(-) & 0.99(1.00) & 0.99(1.00) & 0.98(0.10) & 0.99(1.00)
\\
DeepSight & 0.99(-) & 0.99(1.00) & 0.99(1.00) & 0.65(0.07) & 0.99(1.00)
\\
Flame & 0.99(-) & 0.99(1.00) & 0.99(1.00) & 0.91(0.11) & 0.99(1.00)
\\
MESAS & 0.99(-) & 0.98(1.00) & 0.98(1.00) & 0.70(0.29) & 0.99(1.00)
\\

\hline
\dfh{} & 0.98(-) & 0.99(0.00) & 0.99(0.00) & 0.99(0.00) & 0.99(0.00)
\\
\dfo{} & 0.99(-) & 0.99(0.00) & 0.99(0.00) & 0.98(0.00) & 0.99(0.00)

\end{tabular}%
}
\label{tab:mnist_different_attacks}
\end{table}

\begin{table}[t]
    \centering
    \caption{CA$\uparrow$(ASR$\downarrow$) of FL servers for CIFAR-10 data.}
    \resizebox{\linewidth}{!}{%
        \begin{threeparttable}
            \begin{tabular}{c|c|c|c|c|c|}
                
                & \textbf{No Attack} & \textbf{VTBA}~\cite{gu2019badnets} & \textbf{ITBA}~\cite{chen2017targeted} & \textbf{NBA}~\cite{zhang2022neurotoxin} & \textbf{IBA}~\cite{nguyen2023iba} \\
                \hline
                
                FedAvg & 0.74(-) & 0.64(0.97) & 0.67(1.00) & 0.69(0.73) & 0.73(0.73)
                \\
                DP-SGD & 0.73(-) & 0.67(0.99) & 0.72(1.00) & 0.71(0.81) & 0.72(0.92)
                \\
                {\changed{m-Krum}} & 0.68(-) & 0.72(0.29) & 0.73(0.03) & 0.69(0.97) & 0.73(0.06)
                \\
                FoolsGold & 0.71(-) & 0.69(1.00) & 0.71(0.98) & 0.70(0.71) & 0.70(0.75)
                \\
                DeepSight & 0.72(-) & 0.73(1.00) & 0.70(1.00) & 0.70(0.52) & 0.18(0.99)
                \\
                Flame & 0.72(-) & 0.71(0.89) & 0.71(0.50) & 0.43(0.90) & 0.73(0.42)
                \\
                MESAS$^*$ & 0.72(-) & 0.29(0.13) & 0.64(0.26) & 0.18(0.93) & 0.16(0.29)
                \\
                \hline
                \dfh{} & 0.72(-) & 0.71(0.06) & 0.70(0.02) & 0.71(0.05) & 0.70(0.09)
                \\
                \dfo{} & 0.72(-) & 0.71(0.07) & 0.70(0.02) & 0.71(0.04) & 0.71(0.06)
                \\
            \end{tabular}%
            \begin{tablenotes}
                \item[]$^*$ Under our realistic threat setting, MESAS performs poorly in terms of CA when backdoor clients are present. However, when using the settings from the original paper~\cite{krauss2023mesas}, MESAS achieves good CA.
            \end{tablenotes}
        \end{threeparttable}
    }
    \label{tab:cifar10_different_attacks}
\end{table}

\begin{table}[t]
\centering
\caption{CA$\uparrow$(ASR$\downarrow$) of FL servers for GTSRB dataset.}
\resizebox{\linewidth}{!}{%
\begin{tabular}{c|c|c|c|c|c|}

& \textbf{No Attack} & \textbf{VTBA}~\cite{gu2019badnets} & \textbf{ITBA}~\cite{chen2017targeted} & \textbf{NBA}~\cite{zhang2022neurotoxin} & \textbf{IBA}~\cite{nguyen2023iba} \\
\hline

FedAvg & 0.93(-) & 0.90(1.00) & 0.92(1.00) & 0.89(1.00) & 0.76(0.91)
\\
DP-SGD & 0.86(-) & 0.92(1.00) & 0.92(1.00) & 0.02(0.03) & 0.74(0.93)
\\
{\changed{m-Krum}} & 0.88(-) & 0.91(0.92) & 0.92(1.00) & 0.01(0.01) & 0.88(0.93)
\\
FoolsGold & 0.85(-) & 0.88(1.00) & 0.87(1.00) &  0.02(0.00) & 0.45(0.70)
\\
DeepSight & 0.88(-) & 0.93(1.00) & 0.92(1.00) & 0.90(1.00) & 0.33(0.01)
\\
Flame & 0.87(-) & 0.51(0.46) & 0.42(0.00) & 0.89(1.00) & 0.87(0.99)
\\
MESAS & 0.87(-) & 0.24(0.00) & 0.30(0.00) & 0.09(0.57) & 0.10(0.00)
\\
\hline
\dfh{} & 0.90(-) & 0.91(0.05) & 0.88(0.08) & 0.89(0.00) & 0.88(0.00)
\\
\dfo{} & 0.90(-) & 0.89(0.00) & 0.89(0.00) & 0.90(0.02) & 0.87(0.00)
\\
\end{tabular}%
}
\label{tab:gtsrb_different_attacks}
\end{table}

\subsection{Comparison with the \sota{} defenses} \label{sec:evaluation_ComparisonWithSOTA}
Tables~\ref{tab:mnist_different_attacks}, \ref{tab:cifar10_different_attacks} and \ref{tab:gtsrb_different_attacks} compare the clean accuracy (CA) and ASR of the {\sota{}} backdoor defenses with those of \df{} for MNIST, CIFAR-10 and GTSRB datasets respectively. \df{} shows minimal drop in CA as compared to the baseline (FedAvg, No Attack), and consistently resists different backdoor attacks including the adaptive IBA attack. The performance of \df{} is consistent in both its incarnations---\dfh{} and \dfo{}---across all three datasets used for evaluation.
\dfo{} works on par with \dfh{}, validating our initial hypothesis that backdoored classifiers are indeed adversarially biased and overconfident. 
We attribute this to a better statistical clustering metric (Line~\ref{line:cluster_metric} of Alg~\ref{alg:defense}) that effectively separates clean and backdoored submissions, guided cluster selection enabled by the trust index $\gamma_i$ (Line~\ref{line:cluster_selection} of Alg~\ref{alg:defense}) to identify the best cluster, and stateful filtering of clients the selected cluster (Line~\ref{line:stateful_filtering} of Alg~\ref{alg:defense}) that filters out backdoored submissions occasionally clustered together with the clean submissions.

We observe in Tab.~\ref{tab:mnist_different_attacks}-\ref{tab:gtsrb_different_attacks} that \sota{} defenses can be typically defeated in real-world settings where the clients are randomly sampled even when the rest of the settings are similar.
MESAS~\cite{krauss2023mesas}---a recently proposed defense---fails to achieve sufficient CA under random sampling of clients on CIFAR-10 (Tab.~\ref{tab:cifar10_different_attacks}) and GTSRB (Tab.~\ref{tab:gtsrb_different_attacks}) datasets. Despite that, we argue that this shortcoming of MESAS is also a strength from the defense perspective---when backdoored clients are included in the training, MESAS typically fails to achieve good CA (however, with several exceptions, for example, on the MNIST dataset and on CIFAR-10 dataset against ITBA) thereby, not creating a false sense of trustworthiness. However, when clean clients always outnumber backdoored clients~\cite{krauss2023mesas}, MESAS works well against attacks. This is due to the instability of MESAS, which can be attributed to the multiple interdependent statistical metrics used by MESAS for clustering.
Among the \sota{} defenses reproduced in our experiments, in terms of both CA and ASR, Flame~\cite{nguyen2022flame} and {\changed{m-Krum}}~\cite{fung2018mitigating} give the best performance after \df{} on GTSRB and CIFAR-10 datasets respectively, though in most cases backdoor attacks were successfully inserted into the defended classifier. For example, in Tab.~\ref{tab:cifar10_different_attacks}, {\changed{m-Krum}} achieves $\approx$73\% CA with $~\approx$3\% ASR, but can be backdoored with NBA with $\approx$97\% ASR.

\begin{figure}
    \centering
    \includegraphics[width=0.7\linewidth]{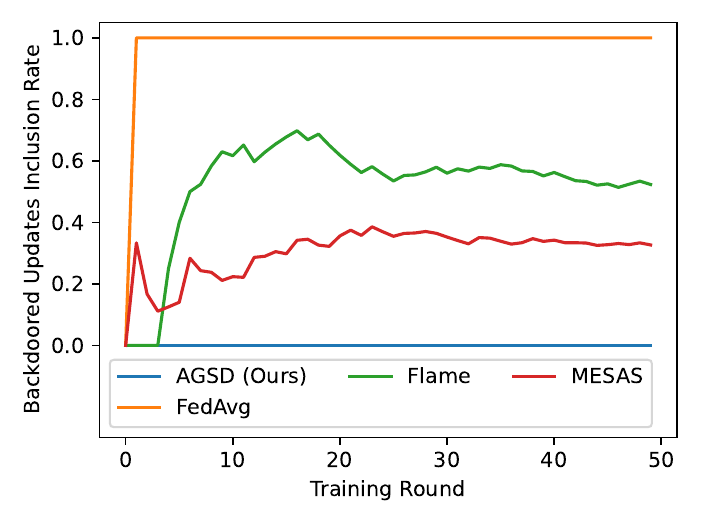}
    \caption{Running average of false negatives (ratio of VTBA clients selected for aggregation) of different servers.}
    \label{fig:number_of_backdoor_clients_selected}
\end{figure}
Despite its stricter selection criteria, \df{} retains high CA on the datasets. In no attack scenario, \df{} only causes a drop in CA from 0\% to 3\% across all datasets. When backdoored clients are present ($\frac{p}{n}=0.45$), \df{} either achieves the best CA or otherwise causes 1\% to 5\% drop as compared to the best CA. Overall, \df{} performs most consistently in terms of both CA and ASR. 
This effectiveness of \df{} is particularly attributed to guided cluster selection and filtering. Fig.~\ref{fig:number_of_backdoor_clients_selected} compares the rate of inclusion of backdoored updates by FedAvg~\cite{mcmahan2017communication}, Flame~\cite{nguyen2022flame}, MESAS~\cite{krauss2023mesas} (two of the most recent FL backdoor defenses) and \df{}. As evident, \df{} identifies backdoored submissions with $\sim$0\% false negatives, unlike Flame and MESAS, which allow backdoored submissions with a significant ratio.
The minimal to no drop in CA is due to the adaptive update mechanism of $\phi_i$ (eq-\eqref{eq:update_trust_history}, which does not significantly penalize (reduction in $\phi_i$) clean clients even if they are rejected in a training round.
\df{} is the only defense that resists the adaptive IBA~\cite{nguyen2023iba}. This is attributed to the better clustering metric explained previously in Fig.~\ref{fig:better_clustering_metric}, as well as the guided cluster selection of \df{}.

\subsection{Robustness of \df{} to FL Hyperparameters}
Here we evaluate the effectiveness of our defense across FL hyperparameters. For this analysis, we train the Resnet-18 classifier on GTSRB (and/or CIFAR-10) datasets for 200 and 500 rounds, respectively, instead of 1000 rounds, and set the patience at 50. All the other parameters and settings stay the same as described in the experimental setup in Section~\ref{subsec:evaluation_experimental_setup}. Specifically, we study the robustness of \df{} to changes in (i) clients sampling ratio $\frac{c}{n} \in \{0.1, 0.2, 0.3, 0.4\}$ for GTSRB dataset, (ii) held-out set size $|D_h| \in \{10, 50, 100, 500, 1000\}$ for GTSRB and CIFAR-10 datasets,
(iii) backdoor clients weight scaling $s_b \in \{1, 2, 3, 5\}$\footnote{It has been observed in the literature that multiplying backdoored updates with $s_b > 1$ improves the attack effectiveness},
\changed{
(iv) the predefined number of clusters for the clustering algorithm in $\{2, 3, 4, 5, 6\}$, and
(v) ratio of backdoored clients in the universal clients' set $\frac{p}{n} \in \{0.01, 0.5, 0.1, 0.15, 0.25, 0.35, 0.45, 0.55, 0.65, 0.75, 0.85\}$.
}
We present results in Appenix, Fig.~\ref{fig:hyperparameter_clients_ratio}, \ref{fig:hyperparameter_healing_set_size}, \ref{fig:hyperparameter_backdoor_scaling}, \ref{fig:hyperparameter_number_clusters}, \ref{fig:backdoor_clients_outnumbering_clean_clients}.
Overall, we observe the \df{} is sufficiently robust to changes in such hyperparameters. This is because the guidance of \df{} inherently opposes backdoor attacks---stronger backdoor attacks show stronger adversarial bias and overconfidence and thus are more easily detected.

We observe that as $\frac{c}{n}$ increases, the CA of the \dfh{} slightly decreases, which is in line with previous observations~\cite{zhu2020empirical}. However, the backdoor ASR in Fig.~\ref{fig:hyperparameter_clients_ratio} remains $~0\%$ irrespective of $\frac{c}{n}$, showing \df{}'s robustness. \df{} is also sufficiently robust to the size of the held-out dataset $|D_h|$ as shown in Fig.~\ref{fig:hyperparameter_healing_set_size} for both GTSRB and CIFAR-10 datasets---\df{} performs well against VTBA even if only 10 samples of $D_h$ are available. This is due to the strength of VTBA, which shows near zero standard deviation in output classes over adversarially perturbed samples. \df{} is also robust to backdoor scaling $s_b$ as illustrated in Fig.~\ref{fig:hyperparameter_backdoor_scaling}, where ASR is $~0\%$ irrespective of $s_b$. This is expected as \df{} uses cosine similarities to cluster and analyze client submissions, which makes \df{} sufficiently agnostic to the backdoor scaling constant. 
\changed{
\df{} is also sufficiently robust to the predefined number of clusters in the clustering algorithm as illustrated in Fig.~\ref{fig:hyperparameter_number_clusters}.
}

Finally, Fig.~\ref{fig:backdoor_clients_outnumbering_clean_clients} shows that \df{} is sensitive to the number of backdoored clients $p$ in the universal clients' set for both GTSRB and CIFAR-10 datasets. Specifically, the CA of \df{} decreases as $\frac{p}{n}$ increases because \df{} only chooses to aggregate clean clients (as they show higher $\gamma_i$ values). Therefore, a decrease in clean clients (or increase in $\frac{p}{n}$) leads to a decrease in the utilizable dataset by \df{}, which in turn results in smaller. On the other hand, ASR of \df{} slightly increases when $p$ is increased to a large value (for example, when $\frac{p}{n}=0.85$ in Fig.~\ref{fig:backdoor_clients_outnumbering_clean_clients}). This is because when $\frac{p}{n}=0.85$, at some training rounds, all the clients sampled by \df{} are backdoored clients, which results in the best cluster also comprising of backdoored clients. Despite that, the ASR of VTBA is less than 15\% because of the stateful filtering stage of \df{}---even if the selected cluster comprises of backdoored clients, they usually have $\phi_i < 0$, and therefore are not used to update the DNN (see the selected cluster in Fig.~\ref{fig:methodology} for illustration).


\begin{table}[t]
\centering
\caption{CA$\uparrow$(ASR$\downarrow$) of \dfh{} and \dfo{} against VTBA with non-IID GTSRB dataset distribution among clients.}
\resizebox{\linewidth}{!}{%
\begin{tabular}{c|ccccc|c|}
& \multicolumn{5}{c|}{\textbf{Standard Non-IID}} & \textbf{MESAS} \\
& $\alpha$=0.1 & $\alpha$=0.3 & $\alpha$=0.5 & $\alpha$=0.7 & $\alpha$=0.9 & \textbf{Non-IID} \\
\hline
FedAvg & 0.93(1.00) & 0.94(1.00) & 0.94(0.99) & 0.91(0.99) & 0.06(0.00) & -
\\
{\changed{m-Krum}} & 0.92(1.00) & 0.93(1.00) & 0.94(1.00) & 0.93(1.00) & 0.06(0.23) & -
\\
Flame & 0.92(1.00) & 0.94(1.00) & 0.94(1.00) & 0.92(0.96) & 0.88(0.99) & -
\\
\hline
\dfh{} & 0.86(0.00) & 0.89(0.00) & 0.87(0.00) & 0.09(0.02) & 0.14(0.00) & 0.89(0.00)
\\
\dfo{} & 0.86(0.00) & 0.88(0.03) & 0.37(0.00) & 0.54(0.45) & 0.07(0.00) & 0.86(0.22)
\\
\end{tabular}%
}
\label{tab:non_iid_evaluation}
\end{table}
\subsection{Evaluation on non-IID Data Distribution}
Following previous works~\cite{wang2020attack, nguyen2022flame}, we study the robustness of \df{} to non-IID data distribution. We consider two types of non-IID data distributions in this experiment: (1) standard non-IID data distribution~\cite{wang2020attack} with varying degrees of non-IID $\alpha$, where $\alpha$ denotes the fraction of images of a specific class in the training dataset assigned to a certain group of clients; and (2) Intra-client non-IID recently identified by MESAS~\cite{krauss2023mesas} as an important and practical non-IID evaluation benchmark. Results in Tab.~\ref{tab:non_iid_evaluation} show that in terms of CA, \sota{} defenses typically work better than \df{} on standard non-IID. However, they achieve this at the cost of being consistently backdoored by VTBA.

Standard non-IID distributions with $\alpha \geq 0.5$ significantly degrade CA of \df{} (Tab.~\ref{tab:non_iid_evaluation}). Although some of the backdoored submissions were able to successfully bypass \df{} defense in several training rounds for non-IID cases, they could not achieve high ASR. However, at $\alpha=0.7$, \dfo{} gets backdoored with 45\% ASR---the highest recorded ASR against \df{}. We observe that the occasional inclusion of backdoored submissions leads to highly unstable training.
When backdoored submissions continue to bypass \df{} once every few training rounds, the training becomes unstable, and CA increases very slowly until the patience of the server runs out (see Fig.~\ref{fig:non_iid5} in Appendix). On the contrary, for MESAS intra-client non-IID data distribution, both \dfh{} and \dfo{} perform well in terms of CA. However, \dfo{} gets backdoored with $~$22\% ASR.

\subsection{Evaluation against Adaptive Attackers}\label{sec:adaptive_evaluation}
We extensively evaluate \df{} under several adaptive backdoor attacks. Specifically, we use three attacks from the current literature: A low-confidence Backdoor Attack (LBA)~\cite{ali2020has}, a recently proposed Multi-Trigger Backdoor Attack (MTBA)~\cite{li2024multi} and Distributed Backdoor Attack (DBA)~\cite{xie2019dba}. We later detail our reasons and intuitions for choosing them for adaptive attack evaluation of \df{}. Moreover, we design two adaptive backdoor attacks keeping in view the defense of \df{}: Adversarially Robust Backdoor Attack (RBA) and Projected Backdoor Attack~(PBA). 
For a comprehensive evaluation, we vary the poisoned data ratio (PDR) $|B_i|/|D_i|$ of adaptive attacks from literature---MTBA, LBA and DBA---and use PDR=0.25 for RBA and PBA. Tab.~\ref{tab:adaptive_evaluation} compares \df{} with {\changed{m-Krum}}~\cite{blanchard2017machine} and Flame~\cite{nguyen2022flame} against adaptive attacks.
\df{} is sufficiently robust to adaptive attacks evaluated in this paper. This is attributed to \df{}'s improved clustering metric, guided cluster selection (\df{} relies on standard deviation of adversarial output classes in addition to their classification confidence), stateful filtering that makes up for occasional errors in the clustering, and the normalized noisy aggregation of client submissions by \df{} that actively regulates backdoor effects caused by rare inclusions of backdoored submissions in the best cluster~\cite{walter2023optimally}.

\begin{table}[t]
\centering
\caption{CA$\uparrow$(ASR$\downarrow$) of \dfh{} and \dfo{} against adaptive backdoor attacks for the GTSRB dataset.}
\resizebox{\linewidth}{!}{%
\color{black}
\begin{tabular}{c|c|ccc|cc|}
& PDR & FedAvg & {\changed{m-Krum}} & Flame & \dfh{} & \dfo{} 
\\ \hline
\multirow{5}{*}{MTBA~\cite{li2024multi}} & 0.25 & 0.89(0.28) & 0.89(0.18) & 0.90(0.22) & 0.85(0.12) & 0.86(0.07)
\\
& 0.35 & 0.91(0.24) & 0.89(0.19) & 0.92(0.28) & 0.86(0.02) & 0.87(0.07)
\\
& 0.45 & 0.91(0.24) & 0.89(0.23) & 0.91(0.21) & 0.87(0.09) & 0.83(0.05)
\\
& 0.55 & 0.34(0.09) & 0.90(0.16) & 0.84(0.12) & 0.87(0.07) & 0.84(0.15)
\\
& 0.65 & 0.91(0.22) & 0.91(0.28) & 0.91(0.04) & 0.85(0.14) & 0.86(0.05)
\\
\hline

\multirow{5}{*}{LBA~\cite{ali2020has}} & 0.25 & 0.90(1.00) & 0.90(0.92) & 0.92(1.00) & 0.91(0.00) & 0.90(0.07)
\\
& 0.35 & 0.91(1.00) & 0.90(1.00) & 0.83(0.91) & 0.90(0.01) & 0.89(0.00)
\\
& 0.45 & 0.89(1.00) & 0.89(0.96) & 0.87(0.60) & 0.78(0.05) & 0.91(0.00)
\\
& 0.55 & 0.89(1.00) & 0.64(0.05) & 0.91(0.92) & 0.90(0.01) & 0.89(0.00)
\\
& 0.65 & 0.89(1.00) & 0.91(1.00) & 0.23(0.00) & 0.81(0.00) & 0.85(0.00)
\\
\hline

\multirow{2}{*}{DBA~\cite{xie2019dba}} & 0.25 & 0.95(1.00) & 0.91(1.00) & 0.92(1.00) & 0.86(0.00) & 0.90(0.05)
\\
& 0.45 & 0.16(0.00) & 0.41(0.00) & 0.25(0.00) & 0.86(0.00) & 0.87(0.00)
\\

\hline \hline
RBA & 0.25 & 0.21(0.20) & 0.87(0.12) & 0.85(0.12) & 0.86(0.00) & 0.67(0.00)
\\
PBA & 0.25 & 0.88(1.00) & 0.88(1.00) & 0.86(1.00) & 0.90(0.01) & 0.90(0.00)
\end{tabular}%
}
\label{tab:adaptive_evaluation}
\end{table}
MTBA inserts backdoors into the classifier for multiple target classes by using different triggers for different targets~\cite{li2024multi}. Therefore, one expects adversarial perturbations to yield multiple classes in eq-\eqref{eq:adv_bias}, thereby increasing the standard deviation of output classes and reducing adversarial bias. Our results in Tab.~\ref{tab:adaptive_evaluation} show that MTBA is relatively weaker than other adaptive attacks. Despite that, \df{} shows the most consistent robustness against MTBA for different values of PDR $\frac{|B_i|}{|D_i|}$.

\changed{
DBA decomposes a trigger into several distinct patterns and distributes them among multiple backdoored clients. Due to the decompsed local triggers, DBA clients show similar updates as the clean clients, which can potentially fool the clustering mechanism of \df{}, thereby making the attack stealthier~\cite{xie2019dba}. However, results in Tab.~\ref{tab:adaptive_evaluation} show that while DBA can successfully backdoor \sota{} defenses (particularly for smaller values of PDR), \df{} is able to consistently resist DBA and outperform \sota{} defenses.
}

LBA uses soft labels to insert backdoors into the DNN~\cite{ali2020has} to regulate the confidence with which backdoored classifiers misclassify poisoned inputs. We chose LBA for evaluation because we expect LBA backdoored classifiers to be less overconfident as compared to VTBA backdoored classifiers. In Tab.~\ref{tab:adaptive_evaluation}, LBA successfully backdoors both {\changed{m-Krum}} and Flame, typically with ~100\% ASR. This is because LBA backdoored submissions are similar to the clean submissions~\cite{ali2020has}, leading to imperfect clustering in \sota{} defenses (see Fig.~\ref{fig:better_clustering_metric}) and are included in the update (see Fig.~\ref{fig:adaptive_illustration} in Appendix). However, larger PDR causes the difference between clean and backdoored clients to increase, allowing \sota{} defenses to distinguish between clean and backdoored submissions, which explains decreased ASR of LBA for higher PDR. On the contrary, we note that \df{} can mostly successfully resist LBA irrespective of the PDR. Two exceptions are \dfo{} (PDR=0.25) and \dfh{} (PDR=0.45), where LBA achieves 7\% and 5\% ASR, respectively.

RBA works by adversarially training the backdoored classifier against the FGSM attack before submitting it to \df{}. The intuition behind RBA is to make backdoored submissions robust to FGSM attacks. This might lead to a larger standard deviation in the output classes of backdoored submissions for adversarially perturbed inputs. Our results in Tab.~\ref{tab:adaptive_evaluation} show that RBA is not very effective against \sota{} defenses (with ~12\% ASR), and completely fails against \df{}.

Given $f_{t-1}$ from the previous round $t-1$, PBA trains two classifiers---a clean classifier $f_{t,i} \gets \mathcal{T}(D_i, f_{t-1})$ and a backdoored classifier $f_{t,i-} \gets \mathcal{T}(D_{i-}, f_{t-1})$---on $D_i$ and $D_{i-}$ respectively for several epochs, and repeatedly projects $f_{t,i-}$ within $l_\infty$ norm of $f_{t,i}$ after each batch-wise training step during an epoch. The $l_\infty$ norm is computed as the median of $\left(f_{t,i}-f_{t-1}\right)$. The intuition behind PBA is to make backdoored submissions similar to the clean submissions and, therefore, cluster them together with the clean submissions for the update so as to fail the clustering algorithms. Although PBA backdoors \sota{} defenses with 100\% ASR in Tab.~\ref{tab:adaptive_evaluation}, our experiments suggest that \df{} can successfully defend against PBA. Again, this is attributed to improved metrics and novel cluster identification mechanisms of \df{}.

\color{black}

\subsection{Discussions}

\begin{figure}
    \centering
    \begin{subfigure}{0.9\linewidth}
        \centering
        \includegraphics[width=0.8\linewidth, page=2]{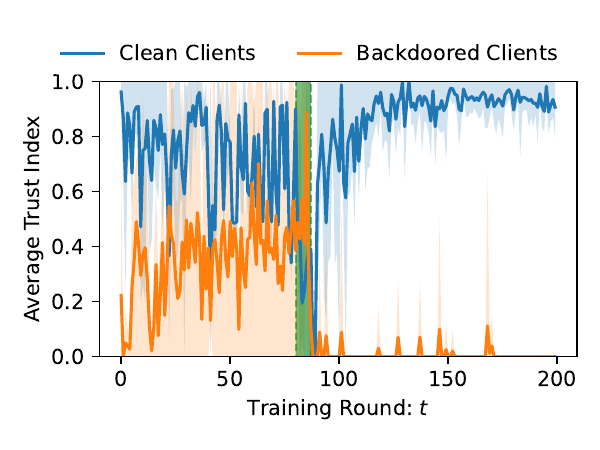}
        \caption{Average trust index $\gamma_i$ of clean and backdoored clients}
    \end{subfigure}
    \begin{subfigure}{0.9\linewidth}
        \centering
        \includegraphics[width=0.8\linewidth, page=3]{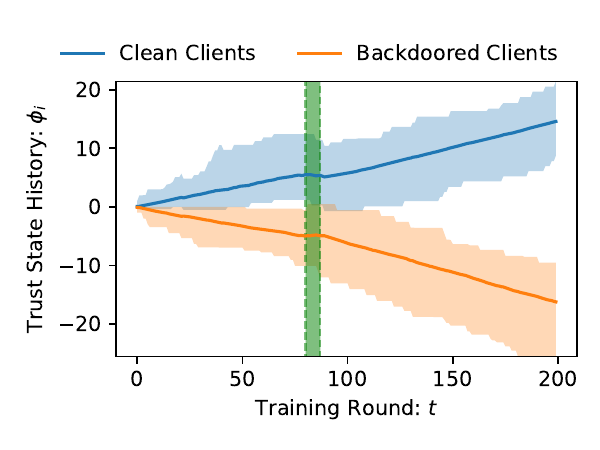}
        \caption{Average trust history $\phi_i$ of clean and backdoored clients}
    \end{subfigure}
    \caption{Comparing the average values of trust index and trust history $\phi_i$ of clean and VTBA backdoored clients sampled in each round $t$ as the training progresses.}
    \label{fig:evolution_of_hgsd_metrics}
\end{figure}
\noindent \textbf{Evolution of trust index and trust history of clients:}
Fig.~\ref{fig:evolution_of_hgsd_metrics}(a) and (b) report average trust index $\gamma_i$ and trust history $\phi_i$ values of clean and VTBA backdoored clients as \dfh{}-defended Resnet-18 trains on GTSRB. Clean submissions, on average, show a notably higher trust index as compared to the backdoored submissions, explaining the effectiveness of \df{}. Fig.~\ref{fig:evolution_of_hgsd_metrics}(b) shows how the backdoored clients are repetitively penalized over time as their trust history devalues. Even if a backdoored client is occasionally clustered among the clean clients, it will ultimately be filtered out of the selected clients due to its bad trust history.

At training round 80 in Fig.~\ref{fig:evolution_of_hgsd_metrics}(a) (green shadowed area), when backdoored submissions show greater $\gamma_i$ compared to clean submissions for a few training rounds, $\phi_i$ of backdoored clients in Fig.~\ref{fig:evolution_of_hgsd_metrics}(b) keeps increasing (green shadowed area) and that of clean clients keeps decreasing. Nevertheless, backdoored submissions are not selected in these rounds because of their negative trust history $\phi_i$. Note that during these rounds, clean submissions are also not selected because of their worse $\gamma_i$.

\begin{figure}
    \centering
    \begin{subfigure}{0.9\linewidth}
        \centering
        \includegraphics[width=0.8\linewidth, page=2]{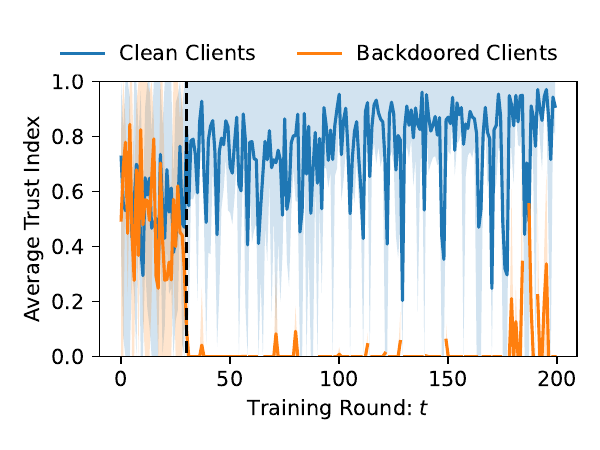}
        \caption{Average trust index $\gamma_i$ of clean and backdoored clients}
    \end{subfigure}
    \begin{subfigure}{0.9\linewidth}
        \centering
        \includegraphics[width=0.8\linewidth, page=3]{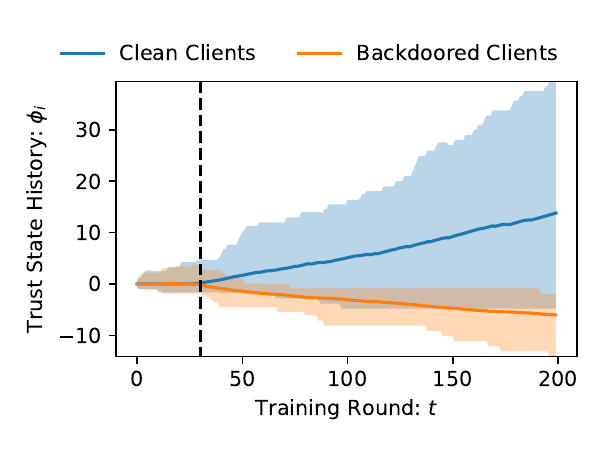}
        \caption{Average trust history $\phi_i$ of clean and backdoored clients}
    \end{subfigure}
    \caption{Average trust index and trust history $\phi_i$ of clean and VTBA backdoored clients who submit clean updates for the initial 30 rounds to develop good trust history.}
    \label{fig:evolution_of_hgsd_metrics_for_nature_changing_clients}
\end{figure}

\noindent 
\textbf{Backdoored clients impersonating clean clients for initial rounds:}
Here, we study the case when backdoored clients intentionally act as clean clients for the initial 30 rounds of training to develop a good first impression (trust history) over \df{} and turn malicious after a few rounds. Fig.~\ref{fig:evolution_of_hgsd_metrics_for_nature_changing_clients}(a) and (b) report the results of Resnet-18 being trained on the GTSRB dataset. It can be observed that for the initial 30 rounds (before the dashed vertical line), \df{} continues accepting the submissions from backdoored clients, and does not devalues their trust history $\phi_i$, for as long as they remain clean. However, as soon as the backdoored clients start submitting backdoored submissions (after the black dashed line in Fig.~\ref{fig:evolution_of_hgsd_metrics_for_nature_changing_clients}), \df{} detects the backdoored submissions because of very small values of $\gamma_i$ and starts degrading their $\phi_i$.

\noindent
\changed{
\textbf{Limitations and future work:}
As shown in Fig.~\ref{fig:time_cost_analysis}, \df{} takes $\sim5\times$ longer to complete a training round for the GTSRB datasets. However, on the MNIST and CIFAR-10 datasets, \df{} completes a round in almost the same time as required by \sota{}. The longer time for GTSRB is attributed to the increased time required for adversarial attacks due to a larger image size of the dataset and greater number of classes, which can be seen as the cost of increased robustness provided by \df{}.
}

\changed{
\df{} uses the FGSM attack to guide the cluster selection due to the computational efficiency of FGSM. Future work should study the compatibility of advanced adversarial attacks (such as PGD, CW, and Auto-attack) with \df{}. However, advanced adversarial attacks will incur additional computation costs, which may not be beneficial considering that FGSM perturbations mostly suffice in guiding the cluster selection. We leave detailed evaluations for future work.
}

\section{Related Work} \label{sec:RelatedWork}
Since their introduction~\cite{gu2019badnets}, several backdoor attacks have been proposed that mainly fall into two categories. Model-agnostic backdoor attacks are independent of the model architecture and training data distribution~\cite{gu2019badnets, chen2017targeted, turner2019label, chen2021badnl}, while model-dependant backdoor attacks assume access to the DNN parameters to serve as an auxiliary knowledge source in order to optimize the stealthiness and efficacy of backdoor attacks~\cite{chen2021deeppoison, cheng2021deep, nguyen2023iba, zhang2022neurotoxin, zhong2020backdoor, xu2022more}. Both of these categories pose a realistic threat to FL, as the clients have access to the updated DNN parameters, making them one of the major concerns of industry practitioners~\cite{kumar2020adversarial}.

Current defenses against backdoor attacks can be broadly categorized into several classes based on their threat setting. Training-time defenses for centralized machine learning (CL) aim to resist backdoor insertion during training on centrally held annotated data~\cite{ali2020has, khaddaj2023rethinking, li2021anti}. These defenses are only applicable to data collection scenarios, assuming that the training data is centrally located on a server.
Our paper falls into the category of training-time defenses for FL where a defending server either (i) identifies and removes backdoored submissions~\cite{blanchard2017machine, rieger2022deepsight, nguyen2022flame, krauss2023mesas, castillo2023fledge}; and/or (ii) robustly aggregates them~\cite{dwork2014algorithmic, fung2018mitigating, walter2023optimally}. Another category of training-time defenses in FL assumes a defending client instead of a defending server~\cite{zhang2023flip}.
Test-time defenses assume that the model has already been backdoored. These defense aim to: (i) detect a backdoored DNN~\cite{kolouri2020universal, wang2020practical}; (ii) detect a poisoned input sample that triggers a backdoored DNN~\cite{gao2019strip}; (iii) invert (regenerate) the backdoor trigger~\cite{zhang2021cassandra, tao2022better}; and (iv) detoxify backdoored DNN~\cite{wu2021adversarial, zeng2021adversarial, wang2019neural, liu2018fine, yan2023dhbe}.

Khaddaj et al.~\cite{khaddaj2023rethinking} note that backdooring features are theoretically indistinguishable from the clean features of training data, implying that a backdoor defense must make implicit assumptions regarding the underlying data distribution to identify and mitigate backdoor attacks as identified by authors in the study~\cite{khaddaj2023rethinking}.
Benign submissions outnumbering clean submissions in every training round is one of the most common assumptions in training-time \sota{} backdoor defenses in FL. We highlight that this assumption is unrealistic since clients are randomly sampled in real-world scenarios and show that when this assumption is invalidated, \sota{} defense struggles to defend against backdoor attacks.

\textbf{Adversarial attacks in backdoor defenses:}
Previous studies have used adversarial perturbations to either invert the backdoor trigger (Cassandra~\cite{zhang2021cassandra}) or detect backdoored DNNs (TrojanNet Detector~\cite{wang2020practical}). TrojanNet Detector (TND) has slight similarity with \df{} in that it uses universal adversarial perturbations (UAP)~\cite{moosavi2017universal} to identify the similarity among the output predictions that are then used to detect backdoored DNNs. However, Cassandra and TND can only be applied after the DNN is strongly backdoored (post-training). In FL, backdoors are inserted gradually even for an undefended DNN, as shown in Fig.~\ref{fig:motivation}. This allows several backdoored submissions to remain undetected by TND, leading to the gradual insertion of backdoors.
Once the similarity is computed, TND uses a threshold to decide whether a DNN is backdoored or not---threshold-based statistical defenses have been shown to fail against low-confidence backdoor attacks~\cite{ali2020has}. Further, because TND~\cite{wang2020practical} uses UAP, it is vulnerable to sample-specific backdoor attacks~\cite{li2021invisible}. \df{} instead uses a novel loss function to compute a different perturbation for each sample and is not reliant on the threshold value to identify backdoored submissions. To attack \df{}, backdoored submissions must meet two conditions: (i) get clustered among the clean submissions; and (ii) show trust index similar to or higher than the clean submissions. This makes it challenging for backdoor attackers to optimize ASR and stealthiness simultaneously.

\section{Conclusions}
Backdoor attacks in realistic FL settings are viewed by industrial AI practitioners as one of the most concerning threats, but state-of-the-art defenses fail to defend against these attacks in realistic settings. In this paper, we highlight and use two properties of backdoored classifiers, adversarial bias and overconfidence, to formulate an Adversarially Guided Stateful Defense (\df{}). \df{} defends FL against backdoor attacks in realistic settings (without making any assumptions regarding the population of sampled clients). We evaluate \df{} on MNIST, CIFAR-10 and GTSRB datasets but focus on GTSRB for detailed analysis due to its practical relevance. We find that \df{} is robust to different \sota{} attacks (including adaptive attacks---intuitively chosen from literature and specifically developed for \df{}) and FL hyperparameters with only a slight drop in clean accuracy. \df{} is notably more sensitive to the degree of non-IID data distribution in terms of CA compared to \sota{} defenses, despite consistently outperforming them in terms of ASR.


\bibliographystyle{IEEEtran}
\bibliography{_bibliography_/biblio_1, _bibliography_/biblio_2, _bibliography_/biblio_3}

\appendix
\begin{algorithm}
    \footnotesize
    \caption{Adversarially Guided Stateful Defense~(\df{}) for the $t^{\text{th}}$ training round}
    \label{alg:defense}
    \begin{algorithmic}[1]
    
        \Input
        \Statex $\{f_{t,i}\}_{\forall i \in S(c, n)}$ all client models sampled in $t^{\text{th}}$ round
        \Statex $\gamma_{t-1, \alpha} \gets$ exponential avg. of previously computed $\gamma_\alpha$ values
        \Statex $\gamma_{t-1, \beta} \gets$ exponential avg. of previously computed $\gamma_\beta$ values
        \Statex $\{\phi_i\}_{i \in S(c, n)} \gets$ the trust index history of all sampled clients
        \Statex $D_h \gets$ small held-out healing dataset available to \df{}
        
        \Output
        \Statex $f_t \gets$ aggregated model of the $t^{\text{th}}$ round
    
        \Statex
        \Procedure{Rescale}{$\Delta=\{\delta_1,..., \delta_i, ..., \delta_c\}$}
            \State $\Delta \gets \left\{ \frac{\delta_i \times \operatorname{median}{ \norm{\Delta}_2 }}{ \norm{\delta_i}_2 } \right\}_{\forall i} $
            \State \textbf{return} $\Delta$
        \EndProcedure
        
        \Statex
        \Procedure{Noisy Aggregate}{$\Delta=\{\delta_1,..., \delta_i, ..., \delta_c\}$}
            \State $\{\sigma_i \gets \operatorname{std}{(\delta_i)}\}_{\forall 1 \leq i \leq c}$
            \State $\Delta_f \gets \frac{1}{z} \times \sum_{\forall 1 \leq i \leq c} \left( \delta_i + 10^{-5} \times \mathcal{N}(0, \sigma_i^2) \right)$;
            \State \textbf{return} $f_{t-1} + \Delta_f$
        \EndProcedure
        
        \Statex
        \Statex \textcolor{gray}{// Preliminary Aggregation and Clustering}
        \State $\Delta^{(s)} \gets \textsc{Rescale} \left( \left\{ f_{t,i}-f_{t-1} \right\}_{i \in S(c, n)} \right)$
        \State $f_{t,-} \gets \textsc{Noisy Aggregate}\left(\Delta^{(s)}\right)$ \textcolor{blue}{\Comment{$f_{t,-}$ is potentially poisoned}}
        \State $\operatorname{cluster\_metric} \gets \textsc{Pair-wise Cos Sim}(f_{t-1}+\Delta^{(s)}-f_{t,-}) + \textsc{Pair-wise Cos Sim}(\Delta^{(s)})$ \label{line:cluster_metric}
        \State $K_1, K_2 \gets \textsc{Cluster}(\operatorname{cluster\_metric})$ \label{line:cosine_of_delta}
        
        \Statex
        \Statex \textcolor{gray}{// Computing the Trust-Index}
        \State \resizebox{1\linewidth}{!}{$D_{adv} \gets \textsc{Adversarial Attack}\left(D_h, \textsc{Fed\_Avg}\left(\left\{f_{t,i}\right\}_{\forall i \in S(c,n)}\right)\right)$}
        \State $p_{i, adv} \gets f_{t,i}(D_{adv})$
        \State $\sigma \gets \{\sigma_i\}_{\forall i} \gets \left\{\underset{D_{adv}}{\operatorname{std}}( \operatorname{onehot}( \operatorname{argmax} p_{i, adv} ) )\right\}_{\forall i}$ \label{alg:std_gamma}
        \State $\eta \gets \{\eta_i\}_{\forall i} \gets \left\{\max\underset{D_{adv}}{\mathbb{E}}\left[p_{i,adv}\right]\right\}_{\forall i \in S(c, n)}$ \label{alg:conf_gamma}
        \State $\sigma \gets \{\sigma_i\}_{\forall i} \gets \{\underset{\forall i}{\operatorname{softmax}}(\sigma_i)\}_{\forall i}$  \label{alg:std_rescaling}
        \State $\eta \gets \{\eta_i\}_{\forall i} \gets \{\underset{\forall i}{\operatorname{softmax}}(\eta_i)\}_{\forall i}$ \label{alg:conf_rescaling}
        \State $\left\{ \gamma_i \gets \sigma_i - e^{\mathcal{W}(\sigma)} \times \eta_i \right\}_{\forall i}$
        \State $\alpha \gets \operatorname{argmax} \left\{ \mathbb{E}[\{\gamma_i\}_{\forall i \in K_1}], \mathbb{E}[\{\gamma_i\}_{\forall i \in K_2}] \right\}$ \label{line:cluster_selection}
        
        \Statex
        \Statex \textcolor{gray}{// Stateful Selection}
        \State $f_t \gets \textsc{Noisy Aggregate}\left(\left\{ f_{t,i}-f_{t-1} \right\}_{\forall i,\ i \in K_\alpha,\ \phi_i > 0}\right)$ \label{line:stateful_filtering}
        \State Update $\phi_i$ with eq-\eqref{eq:update_trust_history}
    
        \Statex
        \State \textbf{return} $f_{t}$ 
    
    \end{algorithmic}
\end{algorithm}

\begin{figure*}[t]
    \centering
    \begin{subfigure}{0.49\linewidth}
        \centering
        \includegraphics[width=1\linewidth]{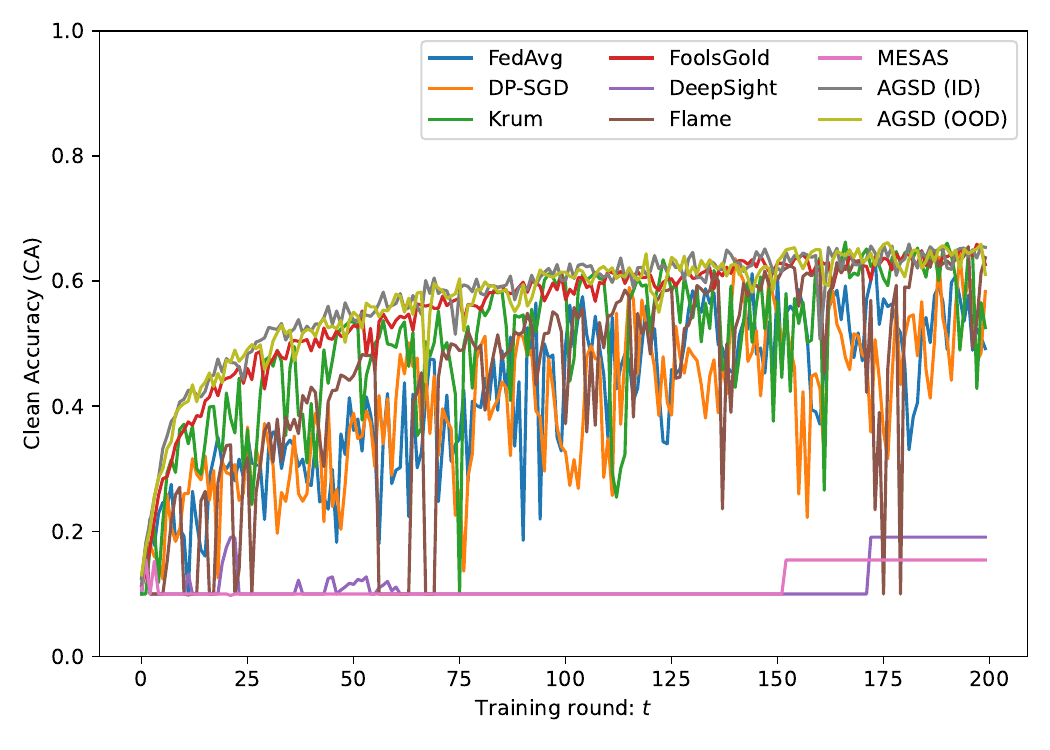}
        \caption{CIFAR-10 dataset}
    \end{subfigure}
    \begin{subfigure}{0.49\linewidth}
        \centering
        \includegraphics[width=1\linewidth]{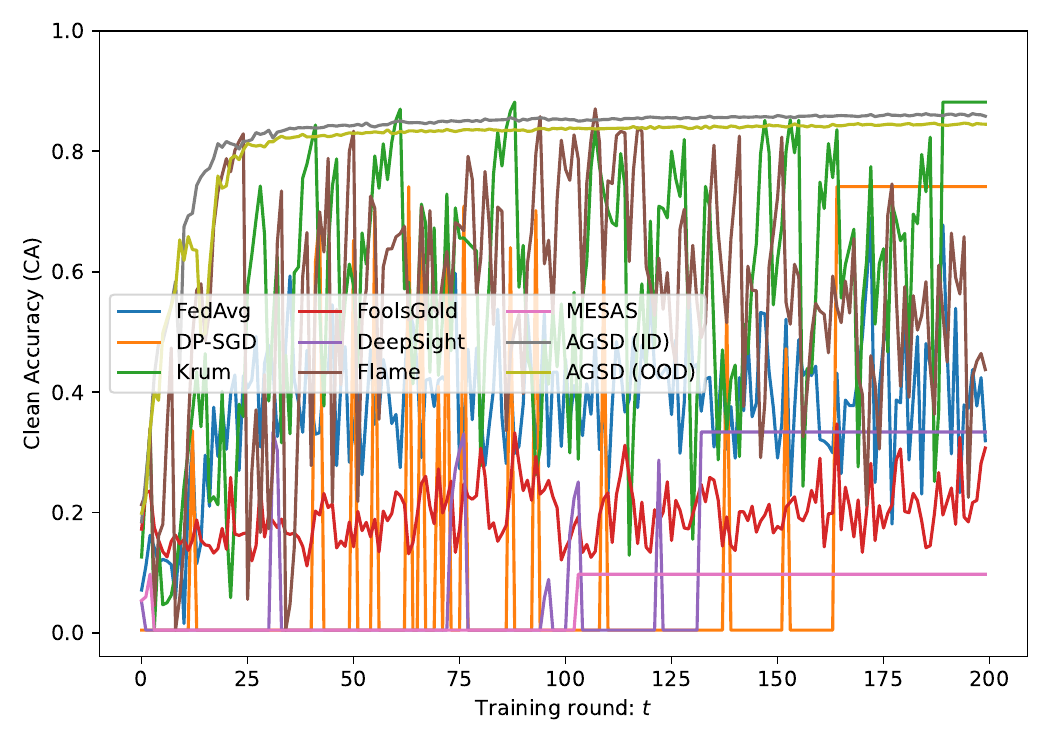}
        \caption{GTSRB dataset}
    \end{subfigure}
    \caption{Comparing the Clean Accuracy (CA) of different servers over the test set for the first 200 training rounds (out of 1000). \take{\df{} is notably faster and more stable than \sota{} defenses because it rejects backdoored submissions that cause unstable training}. \settings{Resnet-18 classifier, number of clients: $n=100$, sample ratio: $\frac{c}{n}=0.1$, 45\% IBA clients.}}
    \label{fig:cifar10_training}
\end{figure*}

\begin{figure}
    \centering
    \includegraphics[width=0.7\linewidth, page=1]{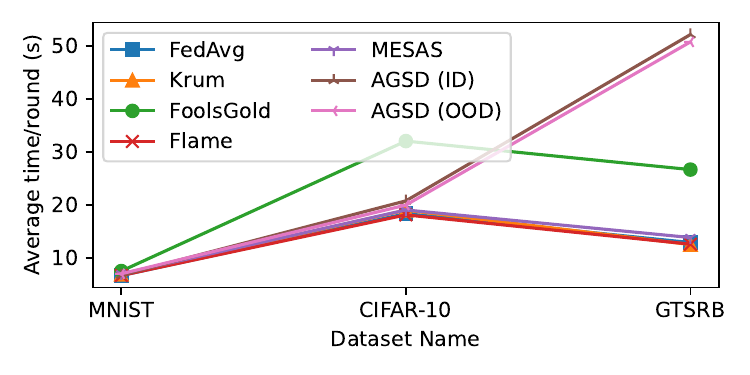}
    \caption{\small
    \changed{
    Reporting average time taken by different FL servers to complete a global round. \take{\df{} completes a round in almost the same time as required by \sota{} FL servers for MNIST and CIFAR-10 datasets. However, for the GTSRB dataset, \df{} takes ~5$\times$ longer to complete a round as the cost of increased robustness. The longer time for GTSRB dataset is attributed to the larger image size of GTSRB datasets and more number of classes which in turn increase the cost of adversarial attack.} \settings{$n=100$, Resnet-18 classifier.}
    }
    }
    \label{fig:time_cost_analysis}
\end{figure}

\begin{figure}
    \centering
    \includegraphics[width=0.7\linewidth, page=1]{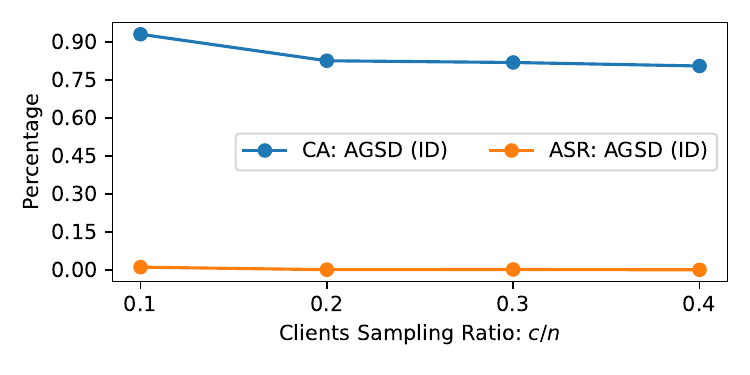}
    \caption{\small Effect of the clients sampling ratio $\frac{c}{n}$ on the clean accuracy (CA) and the backdoor attack success rate (ASR). \take{\df{} (ID) can effectively resist backdoor attacks for different clients sampling ratios.}. \settings{GTSRB dataset and Resnet-18 classifier.}}
    \label{fig:hyperparameter_clients_ratio}
\end{figure}

\begin{figure}[htp]
    \centering
    \begin{subfigure}{1\linewidth}
        \centering
        \includegraphics[width=0.7\linewidth, page=1]{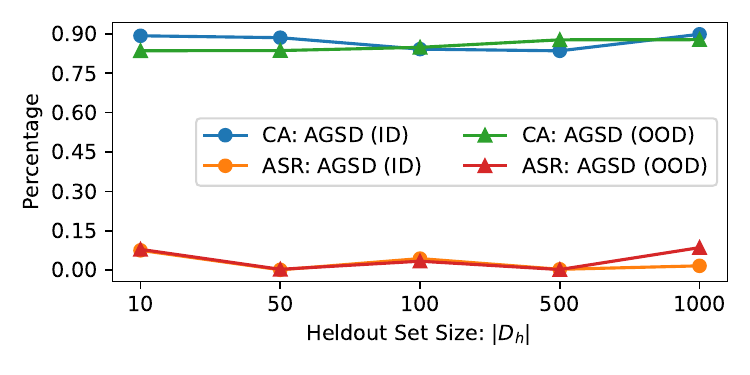}
        \caption{GTSRB dataset}
    \end{subfigure}
    \begin{subfigure}{1\linewidth}
        \centering
        \includegraphics[width=0.7\linewidth, page=2]{figures_agsd/hyperparameter_healing_set_size.pdf}
        \caption{CIFAR-10 dataset}
    \end{subfigure}
    \caption{\small Effect of the held-out set size $D_h$ on the clean accuracy (CA) and the backdoor attack success rate (ASR). \take{\df{} is notably robust to the held-out set size---even a small held-out set of 10 samples can effectively guide the client selection}. \settings{$n=100$, Resnet-18 classifier.}}
    \label{fig:hyperparameter_healing_set_size}
\end{figure}

\begin{figure}[htp]
    \centering
    \includegraphics[width=0.7\linewidth, page=1]{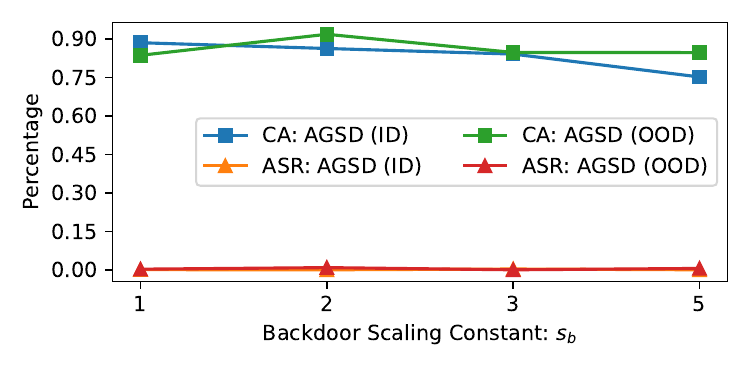}
    \caption{\small Studying the effect of backdoored submissions weight scaling $s_b$ on the robustness of \df{}. \settings{n=100, GTSRB dataset and Resnet-18 classifier.}}
    \label{fig:hyperparameter_backdoor_scaling}
\end{figure}

\begin{figure}[htp]
    \begin{subfigure}{1\linewidth}
        \centering
        \includegraphics[width=0.7\linewidth, page=1]{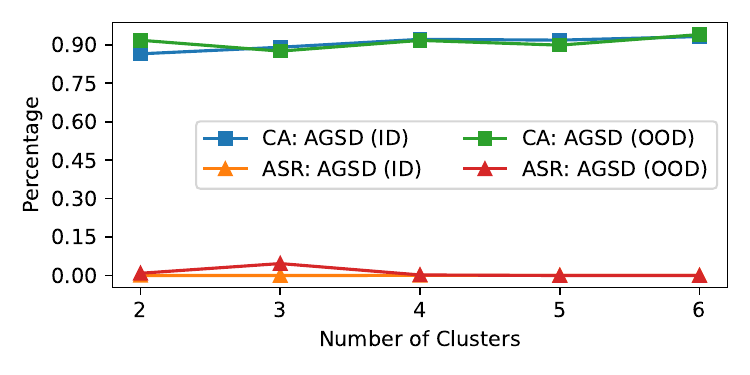}
        \caption{GTSRB dataset}
    \end{subfigure}
    \begin{subfigure}{1\linewidth}
        \centering
        \includegraphics[width=0.7\linewidth, page=2]{figures_agsd/Figure_14_hyperparameter_n_clusters.pdf}
        \caption{CIFAR-10 dataset}
    \end{subfigure}
    \caption{\small
    \changed{
    Studying the effect of the number of clusters on the efficacy of \df{}. \take{\df{} is robust to the number of predefined clusters in terms of both the CA and the backdoor ASR. However, for CIFAR-10 dataset, we observe a very slight increase in the backdoor ASR when number of clusters is large (e.g. 6).} \settings{$n=100$, Resnet-18 classifier.}
    }
    }
    \label{fig:hyperparameter_number_clusters}
\end{figure}

\begin{figure}[htp]
    \begin{subfigure}{1\linewidth}
        \centering
        \includegraphics[width=0.7\linewidth, page=1]{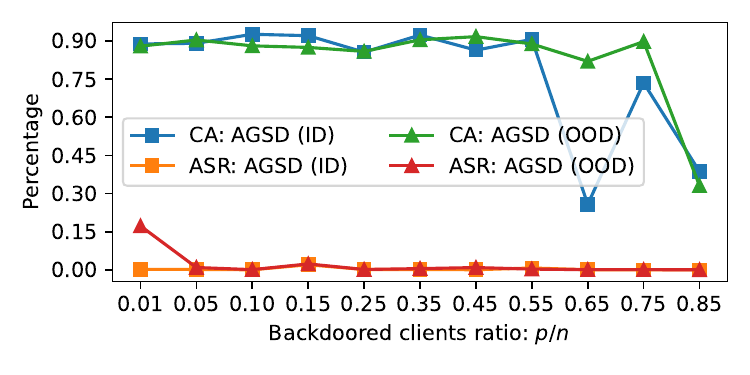}
        \caption{GTSRB dataset}
    \end{subfigure}
    \begin{subfigure}{1\linewidth}
        \centering
        \includegraphics[width=0.7\linewidth, page=2]{figures_agsd/Figure_13_hyperparameter_backdoored_clients_ratio.pdf}
        \caption{CIFAR-10 dataset}
    \end{subfigure}
    \caption{\small
    \changed{
    Studying the effect of $\frac{p}{n}$: the ratio of backdoored clients $p$ to the total number of clients $n$ in the universal clients' set. \take{CA of \df{} decreases as $\frac{p}{n}$ increases. At $\frac{p}{n}=0.65$, CA significantly dropped. We repeated the same experiment multiple times but were not able to reproduce this (small CA) value again. We conjecture that $\frac{p}{n} \geq 0.65$ might cause slightly unstable training rarely leading to small CA of \df{}.} \settings{$n=100$, Resnet-18 classifier.}
    }
    }
    \label{fig:backdoor_clients_outnumbering_clean_clients}
\end{figure}

\begin{figure*}
    \centering
    \begin{subfigure}{0.49\linewidth}
        \centering
        \includegraphics[width=0.8\linewidth, page=2]{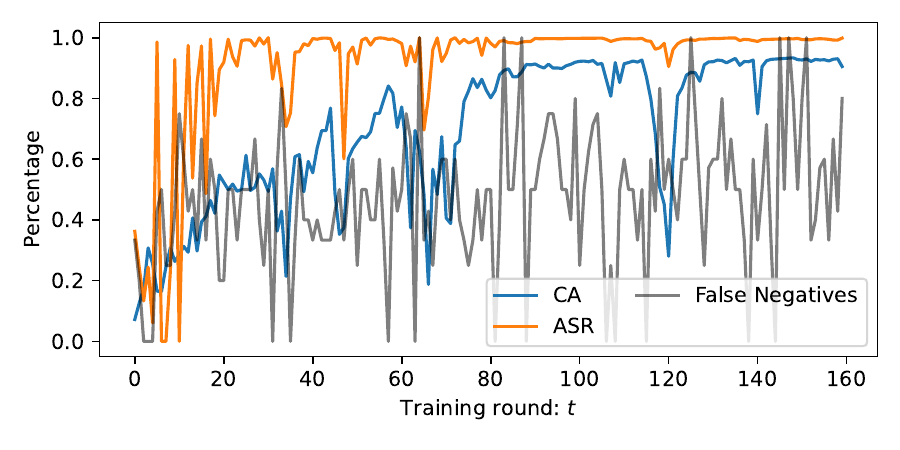}
        \caption{{\changed{m-Krum}}}
    \end{subfigure}
    \begin{subfigure}{0.49\linewidth}
        \centering
        \includegraphics[width=0.8\linewidth, page=3]{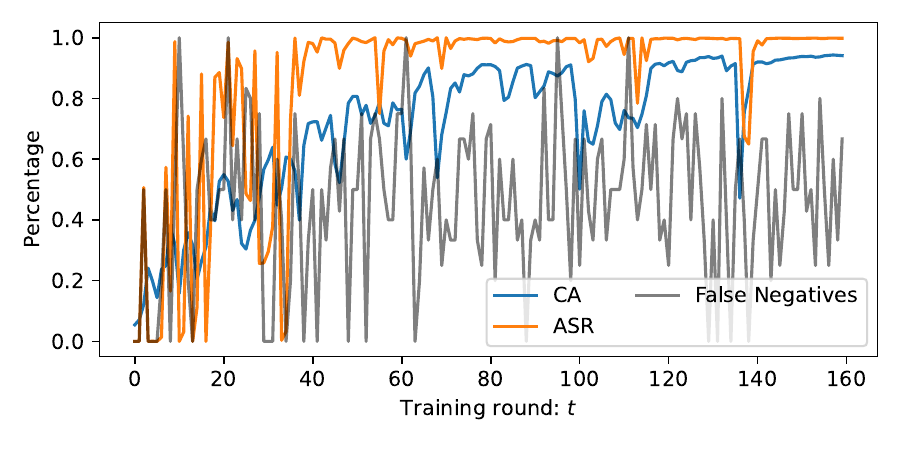}
        \caption{Flame}
    \end{subfigure}
    \begin{subfigure}{0.49\linewidth}
        \centering
        \includegraphics[width=0.8\linewidth, page=4]{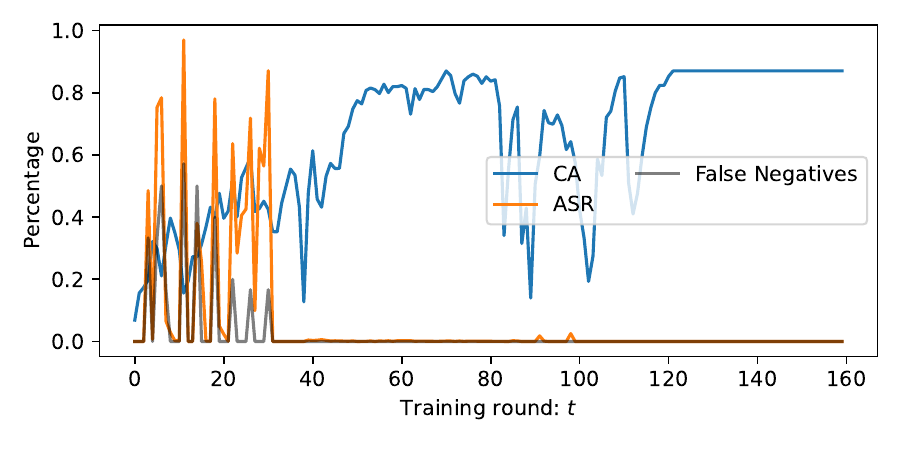}
        \caption{\dfh{}}
    \end{subfigure}
    \begin{subfigure}{0.49\linewidth}
        \centering
        \includegraphics[width=0.8\linewidth, page=5]{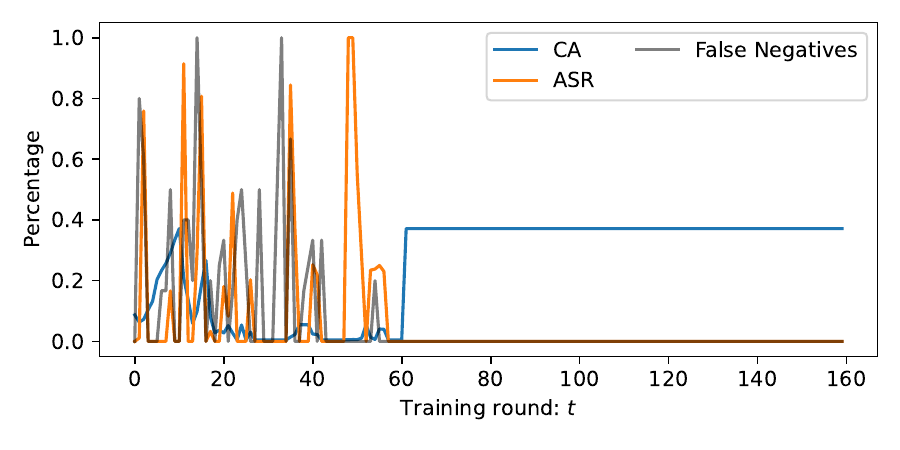}
        \caption{\dfo{}}
    \end{subfigure}
    \caption{Comparing CA, ASR and False Negatives (FN) of different servers for the first few training rounds when dataset is standard non-IID distributed. \take{Figure explains why \df{} is more sensitive to standard non-IID data distribution. Occasional inclusion of backdoored submissions to update clean classifier cause unstable CA. For example, \dfo{} gets backdoored (high FN and ASR at round 38 and CA drops; \dfo{} then resists attack (0 FN) for several rounds until it gets backdoored again at round 55. This continues until the server patience runs out at round 62 and best weights are restored. On contrary, \sota{} defense frequently include backdoored submissions (high FN) resulting in a more stable training}. \settings{GTSRB dataset standard settings, non-IID $\alpha=0.5$}.}
    \label{fig:non_iid5}
\end{figure*}

\begin{figure*}
    \centering
    \begin{subfigure}{0.49\linewidth}
        \centering
        \includegraphics[width=0.8\linewidth, page=2]{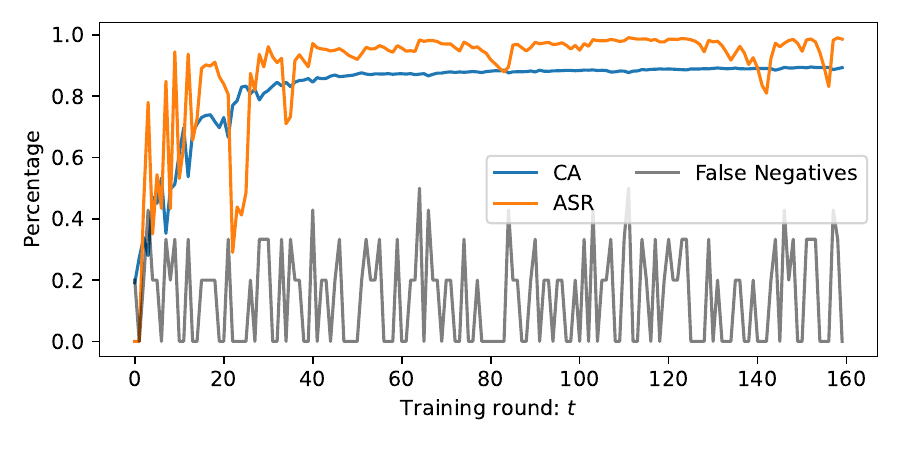}
        \caption{{\changed{m-Krum}}}
    \end{subfigure}
    \begin{subfigure}{0.49\linewidth}
        \centering
        \includegraphics[width=0.8\linewidth, page=3]{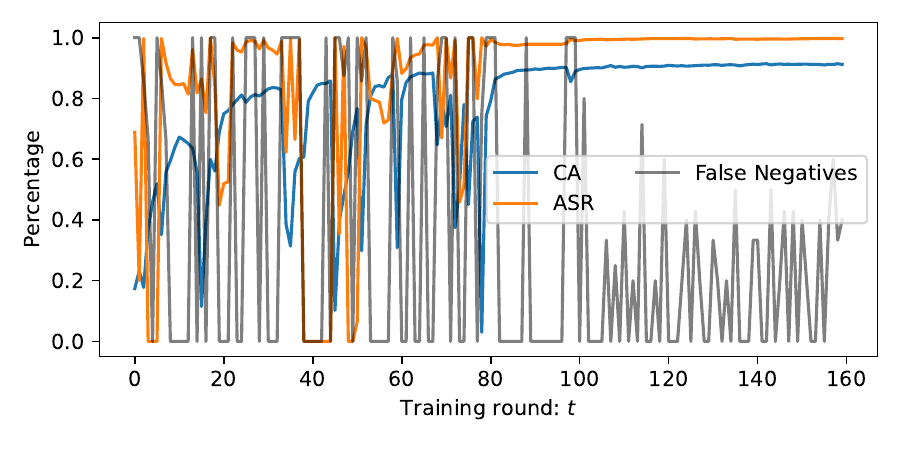}
        \caption{Flame}
    \end{subfigure}
    \begin{subfigure}{0.49\linewidth}
        \centering
        \includegraphics[width=0.8\linewidth, page=4]{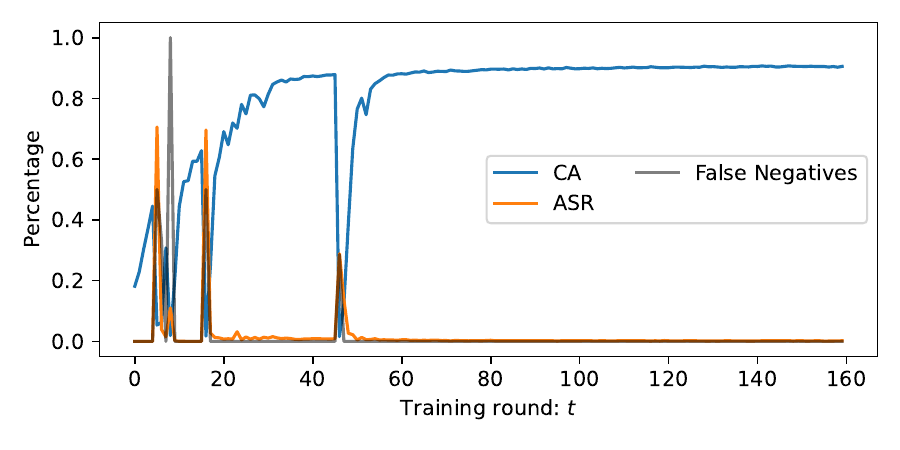}
        \caption{\dfh{}}
    \end{subfigure}
    \begin{subfigure}{0.49\linewidth}
        \centering
        \includegraphics[width=0.8\linewidth, page=5]{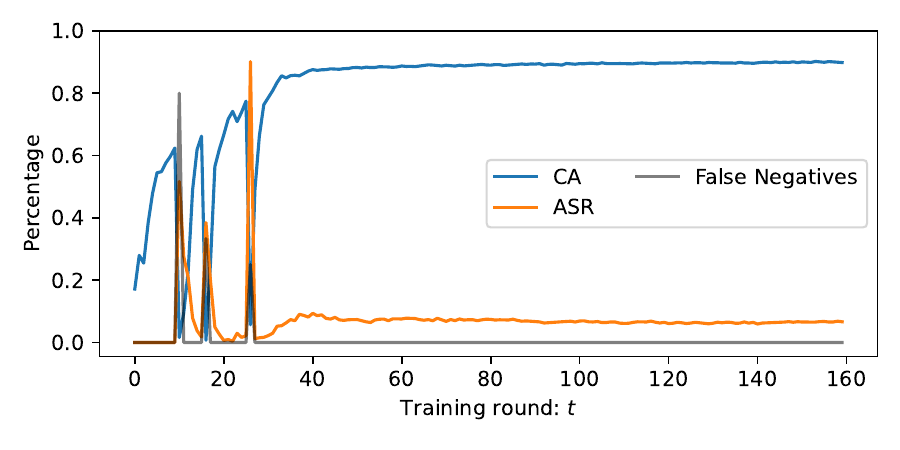}
        \caption{\dfo{}}
    \end{subfigure}
    \caption{Comparing CA, ASR and False Negatives (FN) of different servers for the first few training rounds when against LBA adaptive attack. \take{Again, when backdoored submissions gets passed through \df{}, CA significantly decreases. As compared to \sota{} defenses, \df{} gets bypassed by LBA much fewer times explaining its effectiveness against the adaptive LBA}. \settings{GTSRB dataset standard settings}.}
    \label{fig:adaptive_illustration}
\end{figure*}


\end{document}